\documentclass[10pt,twocolumn,letterpaper]{article}

\usepackage{iccv}
\usepackage{times}
\usepackage{epsfig}
\usepackage{graphicx}
\usepackage{amsmath}
\usepackage{amssymb}
\usepackage{bm}
\usepackage{booktabs}
\usepackage{caption}
\usepackage{comment}
\usepackage{color}

\usepackage{soul}

\usepackage[breaklinks=true,bookmarks=false]{hyperref}

\iccvfinalcopy 

\def\iccvPaperID{1011} 

\begin{document}

\title{P-CNN: Pose-based CNN Features for Action Recognition}

\author{Guilhem Ch\'{e}ron\thanks{
WILLOW project-team, D\'{e}partement d'Informatique de l'\'{E}cole Normale Sup\'{e}rieure, ENS/Inria/CNRS UMR 8548, Paris, France.
}~~\thanks{
LEAR project-team, Inria Grenoble Rh\^{o}ne-Alpes, Laboratoire Jean Kuntzmann, CNRS, Univ. Grenoble Alpes, France.
}\qquad
Ivan Laptev\footnotemark[1]\qquad
Cordelia Schmid\footnotemark[2]\vspace{.3cm}\\
INRIA\vspace{-.0cm}
}

\maketitle

\begin{abstract}
This work targets human action recognition in video.
While recent methods typically represent actions by statistics of local video features, here we argue for the importance of 
a representation derived from human pose.
To this end we propose a new Pose-based Convolutional Neural Network descriptor (P-CNN) for action recognition.
The descriptor aggregates motion and appearance information along tracks of human body parts.
We investigate different schemes of temporal aggregation and experiment with P-CNN features obtained both for automatically estimated and manually annotated human poses. 
We evaluate our method on the recent and challenging JHMDB and MPII Cooking datasets. For both datasets our method shows consistent improvement over the state of the art.


\end{abstract}

\section{Introduction}
Recognition of human actions is an important step toward fully automatic understanding of dynamic scenes. 
Despite significant progress in recent years, action recognition remains a difficult challenge.
Common problems stem from the strong variations of people and scenes in motion and appearance. 
Other factors include subtle differences of fine-grained actions, for example when manipulating
small objects or assessing the quality of sports actions.

The majority of recent methods recognize actions based on statistical representations of local motion descriptors~\cite{LMSR08,Schuldt2004,Wang2013}.
These approaches are very successful in recognizing coarse action (standing up, hand-shaking, dancing) in challenging scenes
with camera motions, occlusions, multiple people, etc. Global approaches, however, are lacking structure and may not be optimal 
to recognize subtle variations, e.g. to distinguish correct and incorrect golf swings or to recognize fine-grained cooking actions illustrated
in Figure~\ref{fig:MPIIqual}.

Fine-grained recognition in static images highlights the importance of spatial structure and spatial alignment as a pre-processing step. 
Examples include alignment of faces for face recognition~\cite{Berg13} as well as alignment of body parts for recognizing species of birds~\cite{Duan12}.
In analogy to this prior work, we believe action recognition will benefit from the spatial and temporal detection and alignment
of human poses in videos.
In fine-grained action recognition, this will, for example, allow to better differentiate \emph{wash hands} from \emph{wash object} actions.

\begin{figure*}
\begin{center}
\includegraphics[trim = 0mm 20mm 95mm 20mm, clip, scale=0.2]{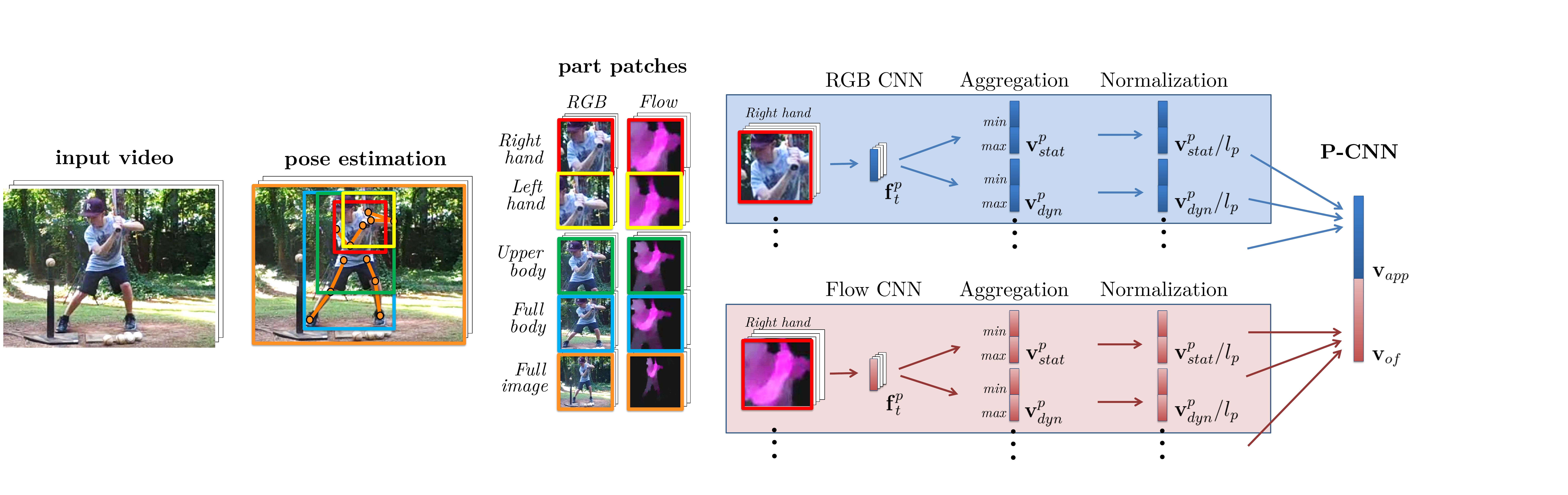}

\end{center}
   \caption{P-CNN features. From left to right:
   Input video. Human pose and corresponding human body parts for one frame of the video. 
   Patches of appearance (RGB) and optical flow for human body parts. 
   One RGB and one flow CNN descriptor $\mathbf{f}_t^p$ is extracted per frame $t$ and per part $p$ (an example is shown for the human body part \emph{right hand}).
   Static frame descriptors $\mathbf{f}_t^p$ are aggregated over time using \emph{min} and \emph{max} to obtain the video descriptor $\mathbf{v}_{stat}^p$.
   Similarly, temporal differences of $\mathbf{f}_t^p$ are aggregated to $\mathbf{v}_{dyn}^p$.
   Video descriptors are normalized and concatenated over parts $p$ and aggregation schemes into appearance features $\mathbf{v}_{app}$ and flow features $\mathbf{v}_{of}$. 
   The final P-CNN feature is the concatenation of $\mathbf{v}_{app}$ and $\mathbf{v}_{of}$.}
\label{fig:pipeline}
\end{figure*}

In this work we design a new action descriptor based on human poses. 
Provided with tracks of body joints over time, our descriptor combines motion and appearance features for body parts.
Given the recent success of Convolutional Neural Networks (CNN)~\cite{Krizhevsky,lecun98}, we explore CNN features obtained separately for each body part in each frame.
We use appearance and motion-based CNN features computed for each track of body parts, and investigate different schemes of temporal aggregation.
The extraction of proposed \emph{Pose-based Convolutional Neural Network} (P-CNN) features is illustrated in Figure~\ref{fig:pipeline}.

Pose estimation in natural images is still a difficult task~\cite{Chen_NIPS14,Tompson14,yang2011articulated}. 
In this paper we investigate P-CNN features both for automatically estimated as well as manually annotated human poses.
We report experimental results for two challenging datasets: JHMDB~\cite{jhuang:hal-00906902}, a subset of HMDB~\cite{Kuehne11} for which manual annotation of human pose have been provided by~\cite{jhuang:hal-00906902}, as well as  MPII Cooking Activities~\cite{rohrbach12cvpr}, composed of a set of fine-grained cooking actions.
Evaluation of our method on both datasets consistently outperforms the human pose-based descriptor HLPF~\cite{jhuang:hal-00906902}.
Combination of our method with Dense trajectory features~\cite{Wang2013} improves the state of the art for both datasets.

The rest of the paper is organized as follows. Related work is discussed in Section~\ref{relatedwork}.
Section~\ref{pcnn} introduces our P-CNN features.
We summarize state-of-the-art methods used and compared to in our experiments in Section~\ref{methods} and present datasets in Section~\ref{dataset}.
Section~\ref{results} evaluates our method and compares it to the state of the art.
Section~\ref{conclusion} concludes the paper.
Our implementation of P-CNN features is available from~\cite{projectwebpage}.

\section{Related work}
\label{relatedwork}

Action recognition in the last decade has been dominated by local features~\cite{LMSR08,Schuldt2004,Wang2013}. In particular, Dense Trajectory (DT) features~\cite{Wang2013} combined with Fisher Vector (FV) aggregation~\cite{perronnin2010improving} have recently shown outstanding results for a number of challenging benchmarks. We use IDT-FV~\cite{Wang2013} (improved version of DT with FV encoding) as a strong baseline and experimentally demonstrate its complementarity to our method.

Recent advances in Convolutional Neural Networks (CNN)~\cite{lecun98} have resulted in significant progress in image classification~\cite{Krizhevsky} and other vision tasks~\cite{Girshick14,Taigman14,toshev2014deeppose}. In particular, the transfer of pre-trained network parameters to problems with limited training data has shown success e.g.~in~\cite{Girshick14,Oquab14,simonyan2014two}. Application of CNNs to action recognition in video, however, has shown only limited improvements so far~\cite{simonyan2014two,Ng15}. We extend previous global CNN methods and address action recognition using CNN descriptors at the local level of human body parts.

Most of the recent methods for action recognition deploy global aggregation of local video descriptors.
Such representations provide invariance to numerous variations in the video but
may fail to capture important spatio-temporal structure.
For fine-grained action recognition, previous methods have represented person-object interactions by joint tracking of hands and objects~\cite{ni2014multiple} or, by linking object proposals~\cite{zhou2015interaction}, followed by feature pooling in selected regions.
%
%
Alternative methods represent actions using positions and temporal evolution of body joints. While reliable human pose estimation is still a challenging task, the recent study~\cite{jhuang:hal-00906902} reports significant gains provided by dynamic human pose features in cases when reliable pose estimation is available. We extend the work~\cite{jhuang:hal-00906902} and design a new CNN-based representation for human actions combining positions, appearance and motion of human body parts.

Our work also builds on methods for human pose estimation in images~\cite{pishchulin2013poselet,modec13,toshev2014deeppose,yang2011articulated} and video sequences~\cite{Cherian14,sapp2011cvpr}. In particular, we build on the method~\cite{Cherian14} and extract temporally-consistent tracks of body joints from video sequences. While our pose estimator is imperfect, we use it to derive CNN-based pose features providing significant improvements for action recognition for two challenging datasets.

%
%

%

\section{P-CNN: Pose-based CNN features} 
\label{pcnn}
We believe that human pose is essential for action recognition. Here, we use positions of body joints to define informative image regions.
We further borrow inspiration from~\cite{simonyan2014two} and represent body regions with motion-based and appearance-based CNN descriptors.
Such descriptors are extracted at each frame and then aggregated over time to form a video descriptor,
see Figure~\ref{fig:pipeline} for an overview. The details are explained below.

To construct P-CNN features, we first compute optical flow~\cite{brox2004high} for each consecutive pair of frames. The method~\cite{brox2004high} has relatively high speed, good accuracy and has been recently used in other flow-based CNN approaches~\cite{actiontubes,simonyan2014two}.
Following~\cite{actiontubes}, the values of the motion field $v_x,v_y$ are transformed to the interval $[0,255]$ by $\tilde v_{x|y} = av_{x|y}+b$ with $a=16$ and $b=128$.
The values below $0$ and above $255$ are truncated. We save the transformed flow maps as images with three channels corresponding to motion $\tilde v_x$, $\tilde v_y$ and the flow magnitude.

Given a video frame and the corresponding positions of body joints,
we crop RGB image patches and flow patches for \emph{right hand}, \emph{left hand}, \emph{upper body}, \emph{full body} and \emph{full image} as illustrated in Figure~\ref{fig:pipeline}.
Each patch is resized to $224 \times 224$ pixels to match the CNN input layer.
To represent appearance and motion patches, we use two distinct CNNs with an architecture similar to~\cite{Krizhevsky}.
Both networks contain 5 convolutional and 3 fully-connected layers.
The output of the second fully-connected layer with $k=4096$ values is used as a frame descriptor ($\mathbf{f}_t^p$).
For RGB patches we use the publicly available ``VGG\nobreakdash-f" network from~\cite{Chatfield14} that has been pre-trained on the ImageNet ILSVRC-2012 challenge dataset~\cite{Deng09imagenet:a}.
For flow patches, we use the motion network provided by~\cite{actiontubes} that has been pre-trained for action recognition task on the UCF101 dataset~\cite{soomro2012ucf101}.

Given descriptors $\mathbf{f}_t^p$ for each part $p$ and each frame $t$ of the video, we then proceed with the aggregation of $\mathbf{f}_t^p$ over all frames to obtain a fixed-length video descriptor.
We consider $min$ and $max$ aggregation by computing minimum and maximum values for each descriptor dimension $i$ over $T$ video frames
\begin{equation}
\label{eq:minmaxaggregation}
\begin{aligned}
m_i &= \min_{1 \le t \le T} \mathbf{f}_t^p(i), \\
M_i &= \max_{1 \le t \le T} \mathbf{f}_t^p(i). \\
\end{aligned}
\end{equation}
The {\em static video descriptor} for part $p$ is defined by the concatenation of time-aggregated frame descriptors as 
\begin{equation}
\label{eq:concatstat}
\mathbf{v}^p_{stat} =  \left[ m_1, ..., m_{k}, M_1, ..., M_{k} \right]^\top.
\end{equation}
To capture temporal evolution of per-frame descriptors, we also consider temporal differences of the form $\Delta \mathbf{f}_t^p =  \mathbf{f}_{t+\Delta t}^p - \mathbf{f}_t^p$ for $\Delta t=4$ frames. Similar to (\ref{eq:minmaxaggregation}) we compute minimum $\Delta m_i$ and maximum $\Delta M_i$ aggregations of $\Delta \mathbf{f}_t^p$ and concatenate them into the {\em dynamic video descriptor}
\begin{equation}
\label{eq:concatdyn}
\mathbf{v}^p_{dyn} =  \left[\Delta m_1, ..., \Delta m_{k}, \Delta M_1, ..., \Delta M_{k} \right]^\top.
\end{equation}
Finally, video descriptors for motion and appearance for all parts and different aggregation schemes are
normalized and concatenated into the P-CNN feature vector. The normalization is performed by dividing video descriptors by the average $L_2$-norm of the $\mathbf{f}_t^p$ from the training set.

In Section~\ref{results} we evaluate the effect of different aggregation schemes as well as the contributions of motion and appearance features for action recognition. 
In particular, we compare ``\emph{Max}'' vs.~``\emph{Max/Min}'' aggregation where ``\emph{Max}'' corresponds to the use of $M_i$ values only while "\emph{Max/Min}" stands for the concatenation of $M_i$ and $m_i$  defined in (\ref{eq:concatstat}) and (\ref{eq:concatdyn}).
\emph{Mean} and \emph{Max} aggregation are widely used methods in CNN video representations.
We choose \emph{Max-aggr}, as it outperforms \emph{Mean-aggr} (see Section~\ref{results}).
We also apply \emph{Min} aggregation, which can be interpreted as a ``non-detection feature".
Additionally, we want to follow the temporal evolution of CNN features in the video by looking at their
dynamics (\emph{Dyn}). Dynamic features are again aggregated using \emph{Min} and \emph{Max}
to preserve their sign keeping the largest negative and positive differences.
The concatenation of static and dynamic descriptors will be denoted by ``\emph{Static+Dyn}''.

The final dimension of our P-CNN is $(5 \times 4 \times 4K) \times 2 =
160K$, i.e., 5 body parts, 4 different aggregation schemes, 4K-dimensional
CNN descriptor for appearance and motion. Note that such a
dimensionality is comparable to the size of Fisher vector~\cite{Chatfield11} used
to encode dense trajectory features~\cite{Wang2013}. P-CNN training is performed using a linear SVM.

\section{State-of-the-art methods}
\label{methods}

In this section we present the state-of-the-art methods used and compared to in our experiments. 
We first present the approach for human pose estimation in videos~\cite{Cherian14} used in our experiments.
We then present state-of-the-art high-level pose features (HLPF)~\cite{jhuang:hal-00906902} and 
improved dense trajectories~\cite{Wang2013}.

\subsection{Pose estimation}
\label{meth_pose}

To compute P-CNN features as well as HLPF features, we need to detect
and track human poses in videos.  We have implemented a video pose estimator
based on~\cite{Cherian14}.  
We first extract poses for individual frames using 
the state-of-the-art approach of Yang and
Ramanan~\cite{yang2011articulated}.
Their approach is based on a deformable part model to
locate positions of body joints (head, elbow, wrist...).  
We re-train their model on the FLIC dataset~\cite{modec13}. 

\begin{figure*}
\begin{center}
\includegraphics[scale=0.42]{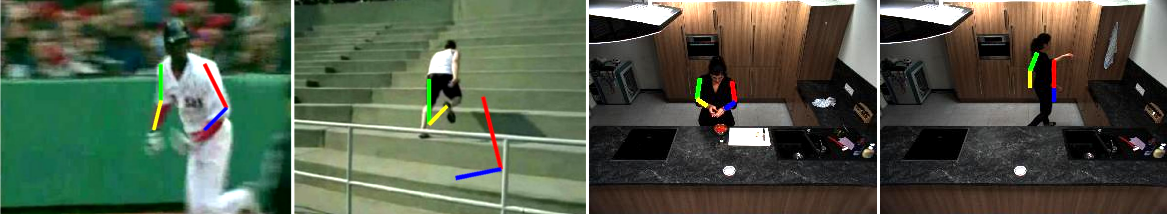}
\end{center}
   \caption{Illustration of human pose estimation used in our experiments~\cite{Cherian14}. Successful
     examples and failure cases on JHMDB (left two images) and on MPII
     Cooking Activities (right two images). Only left and right arms are
     displayed for clarity.} 
\label{fig:poseillustration}
\end{figure*}

Following~\cite{Cherian14}, we extract a large set of pose
configurations in each frame and link them over time using
Dynamic Programming (DP). 
The poses selected with DP are constrained to have a high score of the pose estimator~\cite{yang2011articulated}. 
At the same time, the motion of joints in a pose sequence is constrained to be consistent with the optical flow extracted at joint positions. 
In contrast to~\cite{Cherian14} we do not perform \emph{limb
  recombination}.  See  Figure~\ref{fig:poseillustration} for examples of automatically extracted human poses. 

\subsection{High-level pose features (HLPF)}
\label{sec:posefeatures}
High-level pose features (HLPF) encode spatial and temporal relations of body joint positions
and were introduced in~\cite{jhuang:hal-00906902}.
Given a sequence of human poses $P$,
positions of body joints are first normalized with respect to the person size.
Then, the relative offsets to the head are computed for each pose in $P$.
We have observed that the head is more reliable than the torso used in~\cite{jhuang:hal-00906902}.
Static features are, then, the distances between all pairs of joints, orientations
of the vectors connecting pairs of joints and inner angles spanned by vectors connecting all triplets of joints.

Dynamic features are obtained from trajectories of body joints.
HLPF combines temporal differences of some of the static features, i.e.,
differences in distances between pairs of joints, differences in orientations
of lines connecting joint pairs and differences in inner angles.
Furthermore, translations of joint positions ($dx$ and $dy$) and their orientations ($arctan(\frac{dy}{dx})$) are added.

All features are quantized using a separate codebook for each feature dimension (descriptor type), constructed using $k$-means with $k=20$.
A video sequence is then represented by a histogram of quantized features and the training is performed using an SVM with a $\chi^2$-kernel.
More details can be found in~\cite{jhuang:hal-00906902}.
To compute HLPF features we use the publicly available code with
minor modifications, i.e., we consider the head
instead of the torso center for relative positions. We have also found
that converting angles, originally in degrees, to radians and L2
normalizing the HLPF features improves the performance. 

\subsection{Dense trajectory features}
\label{meth_dt}

Dense Trajectories (DT)~\cite{wang:2011:inria-00583818:1} are local video descriptors that 
have recently shown excellent performance in several action recognition benchmarks~\cite{oneata:hal-00873662,jourdt}. 
The method first densely samples points which are
tracked using optical flow~\cite{Farneback:2003:TME:1763974.1764031}.
For each trajectory, 4 descriptors are computed in the aligned spatio-temporal volume:
HOG~\cite{DT05}, HOF~\cite{LMSR08} and MBH~\cite{Dalal:2006:HDU:2168483.2168522}.
A recent approach~\cite{Wang2013} removes trajectories consistent with the camera motion
(estimated computing a homography using optical flow and SURF~\cite{Bay:2008:SRF:1370312.1370556} point matches and RANSAC~\cite{Fischler:1981:RSC:358669.358692}).
Flow descriptors are then computed from optical flow warped according to the estimated homography. We use the
publicly available implementation~\cite{Wang2013} to compute improved version of DT (IDT).

Fisher Vectors (FV)~\cite{perronnin2010improving} encoding has been shown to outperform
the bag-of-word approach~\cite{Chatfield11} resulting in state-of-the-art performance for action recognition in combination with DT features~\cite{oneata:hal-00873662}.
FV relies on a Gaussian mixture model (GMM) with $K$ Gaussian components,
computing first and second order statistics with respect to the GMM. 
FV encoding is performed separately for the 4 different IDT descriptors (their dimensionality is reduced by the factor of $2$ using PCA).
Following~\cite{perronnin2010improving}, the performance is improved by passing FV through signed square-rooting and $L_2$ normalization.
As in~\cite{oneata:hal-00873662} we use a spatial pyramid representation and a number of $K=256$ Gaussian components.
FV encoding is performed using the Yael library~\cite{douze:hal-01020695} and classification is performed with a linear SVM.

\section{Datasets}
\label{dataset}
\begin{table*}[ht]
\centering 
\renewcommand{\arraystretch}{0.9}
\begin{tabular}{@{}lcccccccccc@{}}\toprule
& &&\multicolumn{3}{c}{\textbf{JHMDB-GT}}   & & &\multicolumn{3}{c}{\textbf{MPII Cooking-Pose\cite{Cherian14}}}\\ \cmidrule{4-6} \cmidrule{9-11} 
 \textbf{Parts}&& &  App   & OF                 & App + OF &  &&  App   & OF & App + OF \\ \midrule
Hands          & &&  $46.3$ & $54.9$             & $57.9$& &  &  $39.9$ & $46.9$ &		$51.9$\\  
Upper body     && &  $52.8$ & $60.9$             & $67.1$&  &  &  $32.3$ & $47.6$ &	$50.1$\\	  
Full body      && &  $52.2$ & $61.6$             & $66.1$& &  &    -    &    -	&	   -\\          
Full image     && &  $43.3$ & $55.7$             & $61.0$& &  &  $28.8$ & $56.2$ & 		$56.5$ \\
All            && &  $\bm{60.4}$ & $\bm{69.1}$  &  $\bm{73.4}$ & & &$\bm{43.6}$ & $\bm{57.4}$ &	$\bm{60.8}$ \\	
\bottomrule 
\end{tabular}
\caption{
 Performance of appearance-based (App) and flow-based (OF) P-CNN features.
 Results are obtained with max-aggregation for JHMDB-GT (\% accuracy) and MPII Cooking Activities-Pose~\cite{Cherian14} (\% mAP).
}
 \label{cnnParts}
\end{table*}

In our experiments we use two datasets JHMDB~\cite{jhuang:hal-00906902} and MPII Cooking Activities~\cite{rohrbach12cvpr},
as well as two subsets of these datasets sub-JHMDB and sub-MPII Cooking. We present them in the following. 

\smallskip

\noindent {\bf JHMDB~\cite{jhuang:hal-00906902}} is a subset of
HMDB~\cite{Kuehne11}, see Figure~\ref{fig:poseillustration} (left). It contains 21 human actions, such as 
\emph{brush hair}, \emph{climb}, \emph{golf}, \emph{run} or
\emph{sit}. Video clips are restricted to the duration of the action. 
There are between 36 and 55 clips per action for a total of
928 clips. Each clip contains between 15 and 40 frames of size $320 \times
240$. Human pose is annotated in each of the 31838 frames. 
There are 3 train/test splits for the JHMDB dataset and evaluation 
averages the results over these three splits. The metric used is
accuracy: each clip is assigned an action label corresponding
to the maximum value among the scores returned by the action classifiers.  

In our experiments we also use a subset of JHMDB, referred to as {\bf
  sub-JHMDB\cite{jhuang:hal-00906902}}. This subset includes 316 clips distributed over 12 actions in
which the human body is fully visible.  Again there are 3 train/test
splits and the evaluation metric is accuracy.

\smallskip 

\noindent {\bf MPII Cooking Activities~\cite{rohrbach12cvpr}} contains
64 fine-grained actions and an additional background class, see
Figure~\ref{fig:poseillustration} (right). 
Actions take place in a kitchen with static background. 
There are 5609 action clips of frame size $1624 \times 1224$. Some
actions are very similar, such as \emph{cut dice}, \emph{cut slices}, and
\emph{cut stripes} or \emph{wash hands} and \emph{wash objects}. Thus, 
these activities are qualified as ``fine-grained''.
There are 7 train/test splits and the evaluation is reported
in terms of mean Average Precision (mAP) using the
code provided with the dataset.

We have also defined a subset of MPII cooking, referred to as {\bf sub-MPII cooking},
with classes  \emph{wash   hands} and \emph{wash
  objects}. We have selected these two classes as they are visually
very similar and differ mainly in manipulated objects.
To analyze the classification performance for these two
classes in detail, we have annotated human pose in all frames of sub-MPII cooking. 
There are  $55$ and $139$ clips for \emph{wash hands} and
\emph{wash objects} actions respectively, for a total of $29,997$ frames.

\section{Experimental results}
\label{results}

This section describes our experimental results and examines the
effect of different design choices. First, we evaluate the complementarity
of different human parts in Section~\ref{partinfo}. We then compare
different  variants for aggregating CNN features  in Section~\ref{cnnvariants}.
Next, we analyze the robustness of our features to errors in the estimated pose and
their ability to classify fine-grained actions in Section~\ref{robustness}.  
Finally, we compare our features to the state of the art and show that
they are complementary to the  popular dense trajectory features in
Section~\ref{featcombination}.

\subsection{Performance of human part features}
\label{partinfo}

Table~\ref{cnnParts} compares the performance of human part
CNN features for both appearance and flow on 
JHMDB-GT (the JHMDB dataset with ground-truth pose) and MPII
Cooking-Pose~\cite{Cherian14} (the MPII Cooking dataset
with pose estimated by~\cite{Cherian14}).
Note, that for MPII Cooking we detect upper-body poses only since full bodies are
not visible in most of the frames in this dataset.

Conclusions for both datasets are similar. We can observe that all
human parts (hands, upper body, full body) as well as the full image
have similar performance and that their combination improves the
performance significantly. Removing one part at a time from this
combination results in the drop of performance (results not shown here).
We therefore use all pose parts together with the full image descriptor
in the following evaluation.
We can also observe that flow descriptors consistently outperform appearance descriptors by a 
significant margin for all parts as well as for the overall combination
{\it All}. Furthermore, we can observe that the combination of
appearance and flow further improves the performance for all parts
including their combination {\it All}.  
This is the pose representation used in the rest of the evaluation.

In this section, we have applied the max-aggregation (see
Section~\ref{pcnn}) for aggregating features extracted for individual
frames into a video descriptor. Different aggregation schemes will be
compared in the next section.

\subsection{Aggregating P-CNN features}
\label{cnnvariants}

CNN features $\mathbf{f}_t$ are first extracted for each frame and
the following temporal aggregation pools feature values for each 
feature dimension over time (see  Figure~\ref{fig:pipeline}).
Results of max-aggregation for JHMDB-GT are reported in
Table~\ref{cnnParts} and compared with other aggregation schemes in Table~\ref{cnnvar}.
Table~\ref{cnnvar}
shows the impact of adding min-aggregation (\emph{Max/Min-aggr}) and the first-order difference between CNN features (\emph{All-Dyn}).
Combining per-frame CNN features and their first-order differences
using max- and min-aggregation further improves results.
Overall, we obtain the best results with \emph{All-(Static+Dyn)(Max/Min-aggr)} for App + OF, 
i.e., $74.6\%$ accuracy on JHMDB-GT. This represents an improvement over \emph{Max-aggr} by $1.2\%$.
On MPII Cooking-Pose~\cite{Cherian14} this version of P-CNN
achieves $62.3\%$ mAP (as reported in
Table~\ref{pose2}) leading to an $1.5\%$ improvement over
max-aggregation reported in Table~\ref{cnnParts}.

\begin{table}[htbp]\centering
\renewcommand{\arraystretch}{0.9}
 \begin{tabular}{@{}lccc@{}}\toprule
\textbf{Aggregation scheme}& \textbf{App}  & \textbf{OF} & \textbf{App+OF} \\ \midrule
All(Max-aggr)                 &  $60.4$ & $69.1$ &	$73.4$     \\ 
All(Max/Min-aggr)             &  $60.6$ & $68.9$ &    $73.1$    		  \\ 
All(Static+Dyn)(Max-aggr)     &  $62.4$ & $\bm{70.6$} &   $74.1$        \\ 
All(Static+Dyn)(Max/Min-aggr) &  $\bm{62.5}$ & $70.2$ &   $\bm{74.6}$        \\ 
All(Mean-aggr)                &  $57.5$ & $69.0$      &   $69.4$        		  \\
\bottomrule 
 \end{tabular}
  \caption{Comparison of different aggregation schemes: \emph{Max}-, \emph{Mean}-,
    and \emph{Max/Min}-aggregations as well as adding first-order
    differences (\emph{Dyn}). Results are given for
    appearance~(\emph{App}), optical flow (\emph{OF}) and App + OF on
    JHMDB-GT (\% accuracy). 
}
 \label{cnnvar}
\end{table}

~\vspace{-0.5cm}\\
We have also experimented with second-order differences and other
statistics, such as mean-aggregation (last row in Table~\ref{cnnvar}),
but this did not improve results. Furthermore, we have tried temporal aggregation of classification
scores obtained for individual frames.  This led to a decrease of
performance, \eg for \emph{All (App)} on JHMDB-GT
score-max-aggregation results in $56.1\%$ accuracy, compared to 
$60.4\%$ for features-max-aggregation (top row, left column in
Table~\ref{cnnvar}). This indicates that early aggregation
works significantly better in our setting. 

In summary, the best performance is obtained for \emph{Max-aggr} on single-frame features,
if only one aggregation scheme is used.
Addition of \emph{Min-aggr} and first order differences \emph{Dyn}
provides further improvement.
In the remaining evaluation we report results for this version of
P-CNN, i.e., \emph{All} parts \emph{App+OF} with \emph{(Static+Dyn)(Max/Min-aggr)}.

\subsection{Robustness of pose-based features}
\label{robustness}

This section examines the robustness of P-CNN features in the
presence of pose estimation errors and compares results with
the state-of-the-art pose features HLPF~\cite{jhuang:hal-00906902}. 
We report results using the code of~\cite{jhuang:hal-00906902} with
minor modifications described in Section~\ref{sec:posefeatures}.   
Our HLPF results are comparable to~\cite{jhuang:hal-00906902} in
general and  are slightly better on JHMDB-GT ($77.8\%$ vs.~$76.0\%$).
%
Table~\ref{pose1} evaluates the impact of automatic pose estimation
versus ground-truth pose (GT)
for sub-JHMDB and JHMDB. We can
observe that results for GT pose are comparable on both datasets and for both type of pose features.
However,  P-CNN is significantly more robust to errors in pose estimation. For automatically estimated poses
P-CNN drops only by $5.7\%$ on sub-JHMDB and by $13.5\%$ on JHMDB, whereas HLPF 
drops by $13.5\%$ and $52.5\%$ respectively. For both
descriptors the drop is less significant on sub-JHMDB, as this subset
only contains full human poses for which pose is easier to estimate.
Overall the performance of P-CNN features for automatically extracted poses
is excellent and outperforms HLPF by a very large margin
($+35.8\%$) on JHMDB.

\begin{table}[htbp]\centering
\renewcommand{\arraystretch}{0.9}
 \begin{tabular}{@{}lccccc@{}}\toprule
                         	&                           \multicolumn{5}{c}{\textbf{sub-JHMDB}}                        \\ \cmidrule{2-6}
                                & GT   &     &   Pose~\cite{yang2011articulated}  & & Diff \\ \midrule
	P-CNN 	                &      $72.5$                      &     &      $66.8$  && $5.7$                    \\
	HLPF	      		&        $78.2$                    &     &      $51.1$  && $27.1$            \\ 
	\bottomrule 
 \end{tabular}
~\vspace{0.3cm}\\
~\\
 \begin{tabular}{@{}lccccc@{}}\toprule
                         	&                           \multicolumn{5}{c}{\textbf{JHMDB}}                        \\ \cmidrule{2-6}
                                & GT   &     &   Pose~\cite{Cherian14}  & & Diff \\ \midrule
	P-CNN 	                &      $74.6$                   &     &      $61.1$  && $13.5$                    \\
	HLPF	      		&        $77.8$                    &     &      $25.3$  && $52.5$            \\ 
	\bottomrule 
 \end{tabular}
  \caption{Impact of automatic pose estimation versus ground-truth pose (GT) for P-CNN features and HLPF~\cite{jhuang:hal-00906902}.
Results are presented for sub-JHMDB and JHMDB (\% accuracy).}

 \label{pose1}
\end{table}

We now compare and evaluate the robustness of P-CNN and HLPF features on the MPII cooking
dataset. To evaluate the impact of ground-truth pose (GT), we have
manually annotated two classes ``washing hand'' and ``washing
objects'', referred to by sub-MPII 
Cooking.  Table~\ref{pose2} compares P-CNN and HLPF for sub-MPII and 
MPII Cooking. In all cases P-CNN outperforms HLPF significantly.
Interestingly, even for ground-truth poses P-CNN performs significantly
better than HLPF.
This could be explained by the better encoding of image appearance by
P-CNN features, especially for object-centered actions such as ``washing hands'' and ``washing objects''.
We can also observe that the drop due to errors in pose estimation is
similar for P-CNN and HLPF. This might be explained by the fact that 
CNN hand features are quite sensitive to the pose estimation. 
However, P-CNN still significantly outperforms HLPF for
automatic pose. In particular, there is a significant gain of $+29.7\%$ for the
full MPII Cooking dataset.

\begin{table}\centering
\renewcommand{\arraystretch}{0.9}
 \begin{tabular}{@{}lccccc@{}}\toprule
                         	&
                                \multicolumn{5}{c}{\textbf{sub-MPII Cooking}}                        \\ \cmidrule{2-6}
                                &      GT          &     &   Pose~\cite{Cherian14}  & & Diff \\ \midrule
	P-CNN 	                &      $83.6$      &     &      $67.5$  && $16.1$                    \\
	HLPF	      		&      $76.2$      &     &      $57.4$  && $18.8$            \\ 
	\bottomrule 
 \end{tabular}
~\vspace{0.3cm}\\
~\\
 \begin{tabular}{@{}lcc@{}}\toprule
                         	&
                                \multicolumn{1}{c}{\textbf{MPII Cooking}}                        \\ \cmidrule{2-2}
                                &     Pose~\cite{Cherian14}  \\ \midrule
	P-CNN 	                &      $62.3$                                     \\
	HLPF	      		&        $32.6$                               \\ 
	\bottomrule 
 \end{tabular}
  \caption{Impact of automatic pose estimation versus ground-truth pose (GT) for P-CNN features and HLPF~\cite{jhuang:hal-00906902}.
Results are presented for sub-MPII Cooking and MPII Cooking (\% mAP).}

 \label{pose2}
\end{table}

\subsection{Comparison to the state of the art}
\label{featcombination}

In this section we compare to state-of-the-art dense
trajectory features~\cite{Wang2013} encoded by
Fisher vectors~\cite{oneata:hal-00873662} (IDT-FV), briefly described
in Section~\ref{meth_dt}. We use the online available code, which we  
validated on Hollywood2 ($65.3\%$ versus $64.3\%$~\cite{Wang2013}).
Furthermore, we show that  our pose features P-CNN and IDT-FV are
complementary and compare to other state-of-the-art approaches on JHMDB
and MPII Cooking.    
 
\begin{figure}
\begin{center}
\includegraphics[trim = 4mm 243mm 155mm 0mm, clip, scale=1.4]{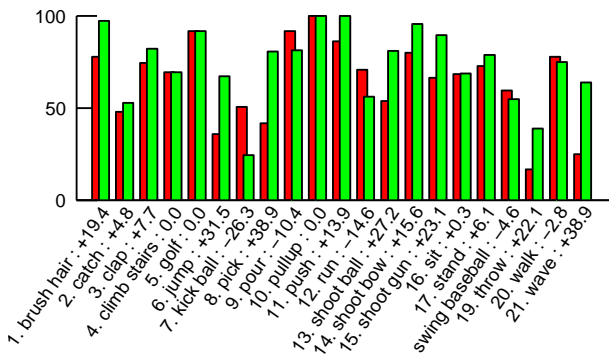}
\end{center}
   \caption{Per class accuracy on JHMDB-GT for P-CNN (green) and IDT-FV (red) methods.
   Values correspond to the difference in accuracy between P-CNN and IDT-FV
   (positive values indicate better performance of P-CNN).}
\label{fig:JHMDBbar}
\end{figure}

Table~\ref{featcomres} shows that for ground-truth
poses our P-CNN features outperform state-of-the-art IDT-FV
descriptors significantly ($8.7\%$). If the pose is extracted
automatically both methods are on par. Furthermore, in all cases the
combination of P-CNN and IDT-FV obtained by late fusion of the
individual classification scores significantly
increases the performance over using individual features only.
Figure~\ref{fig:JHMDBbar} illustrates per-class results for P-CNN and IDT-FV on JHMDB-GT.

\begin{table}\centering
\renewcommand{\arraystretch}{0.9}
 \begin{tabular}{@{}lcccc@{}}\toprule
 	 					& \multicolumn{2}{c}{\textbf{JHMDB}}	&& \textbf{MPII Cook.}	\\  \cmidrule{2-3} \cmidrule{5-5}
	\textbf{Method}     	&	   GT &Pose~\cite{Cherian14}		&&  Pose~\cite{Cherian14}\\  \midrule
	P-CNN 	            	& 	    $74.6$    		&   $61.1$  	&&	$62.3$        		\\  
	IDT-FV 	            	& 	    $65.9$    		&   $65.9$  	&& 	$67.6$       		\\  \midrule
	P-CNN + IDT-FV 	    	& 	    $\bm{79.5}$    		&$\bm{72.2}$	&& 	$\bm{71.4}$     		\\ 
\bottomrule 	
 \end{tabular}
  \caption{Comparison to IDT-FV on JHMDB (\% accuracy) and
    MPII Cooking Activities (\% mAP) for ground-truth (GT) and pose~\cite{Cherian14}.
The combination of P-CNN + IDT-FV performs best.
}
 \label{featcomres}
\end{table}

Table~\ref{globresults} compares our results to other methods 
on MPII Cooking. Our approach outperforms the state of the art
on this dataset and is on par with the recently published work of~\cite{zhou2015interaction}.
We have compared our method with HLPF~\cite{jhuang:hal-00906902} on JHMDB in the previous section.
P-CNN perform 
on par with HLPF for GT poses and significantly outperforms HLPF for automatically estimated poses. 
Combination of P-CNN with IDT-FV improves the performance to $79.5\%$ and $72.2\%$ for GT and automatically estimated poses respectively (see Table~\ref{featcomres}).
This outperforms the state-of-the-art result reported in~\cite{jhuang:hal-00906902}.

\begin{figure*}
\includegraphics[trim = 54mm 37mm 36mm 26mm, clip,scale=0.10]{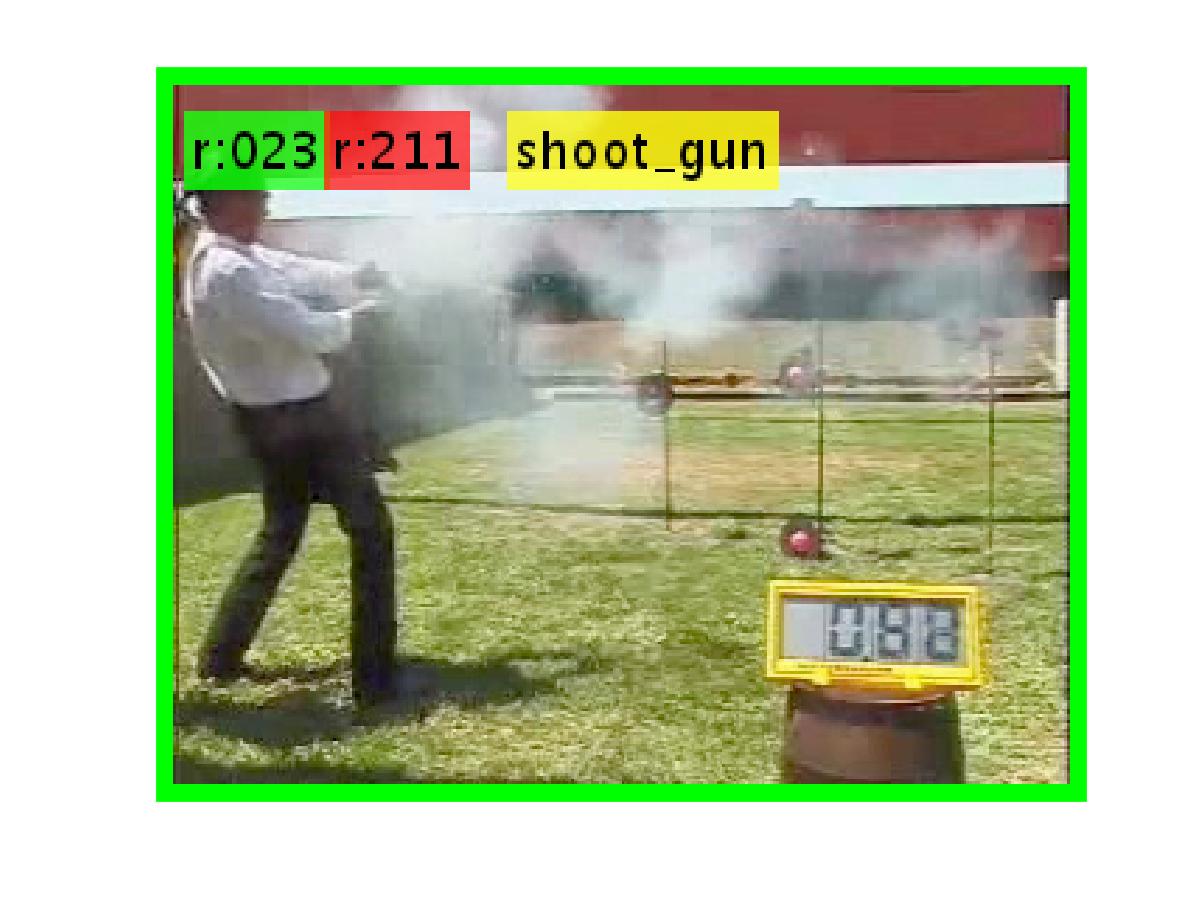}
\includegraphics[trim = 54mm 37mm 36mm 26mm, clip,scale=0.10]{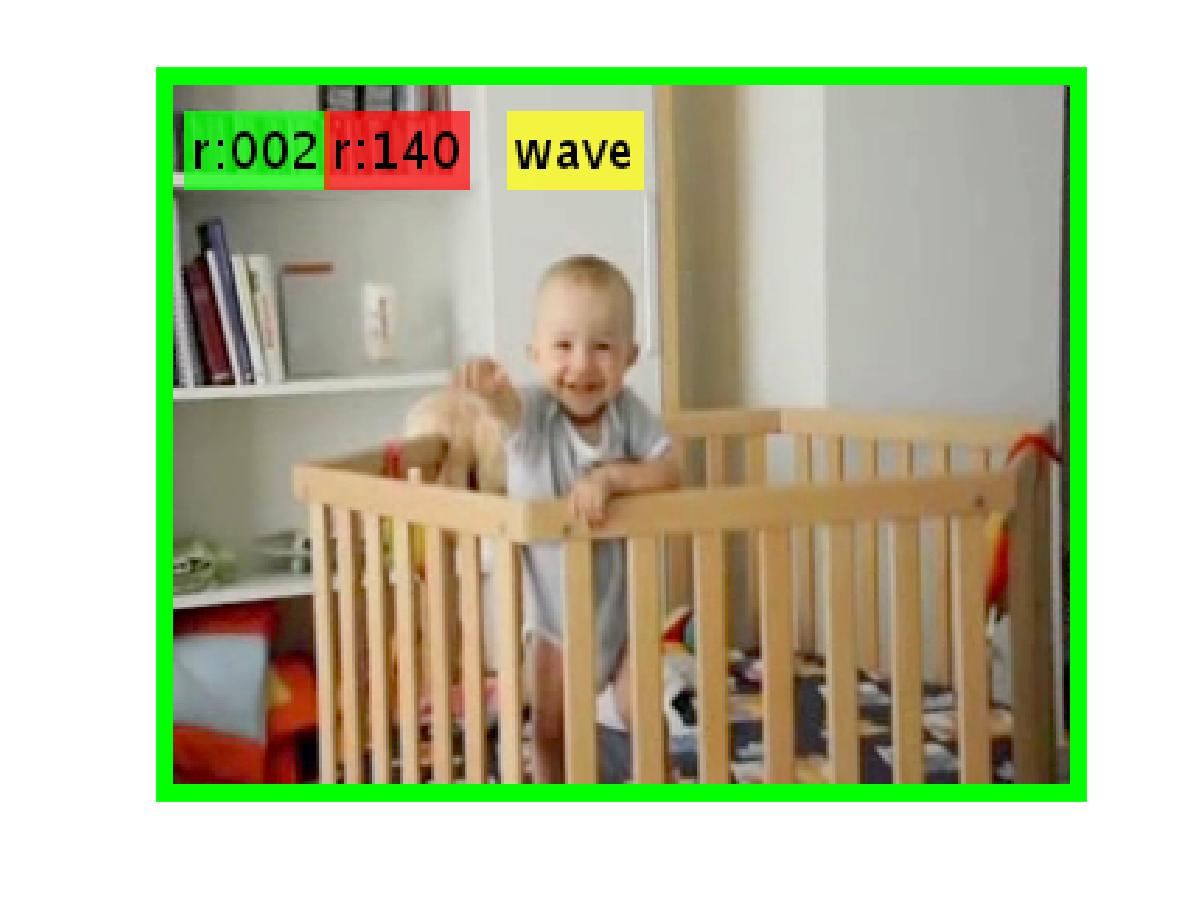}
\includegraphics[trim = 54mm 37mm 36mm 26mm, clip,scale=0.10]{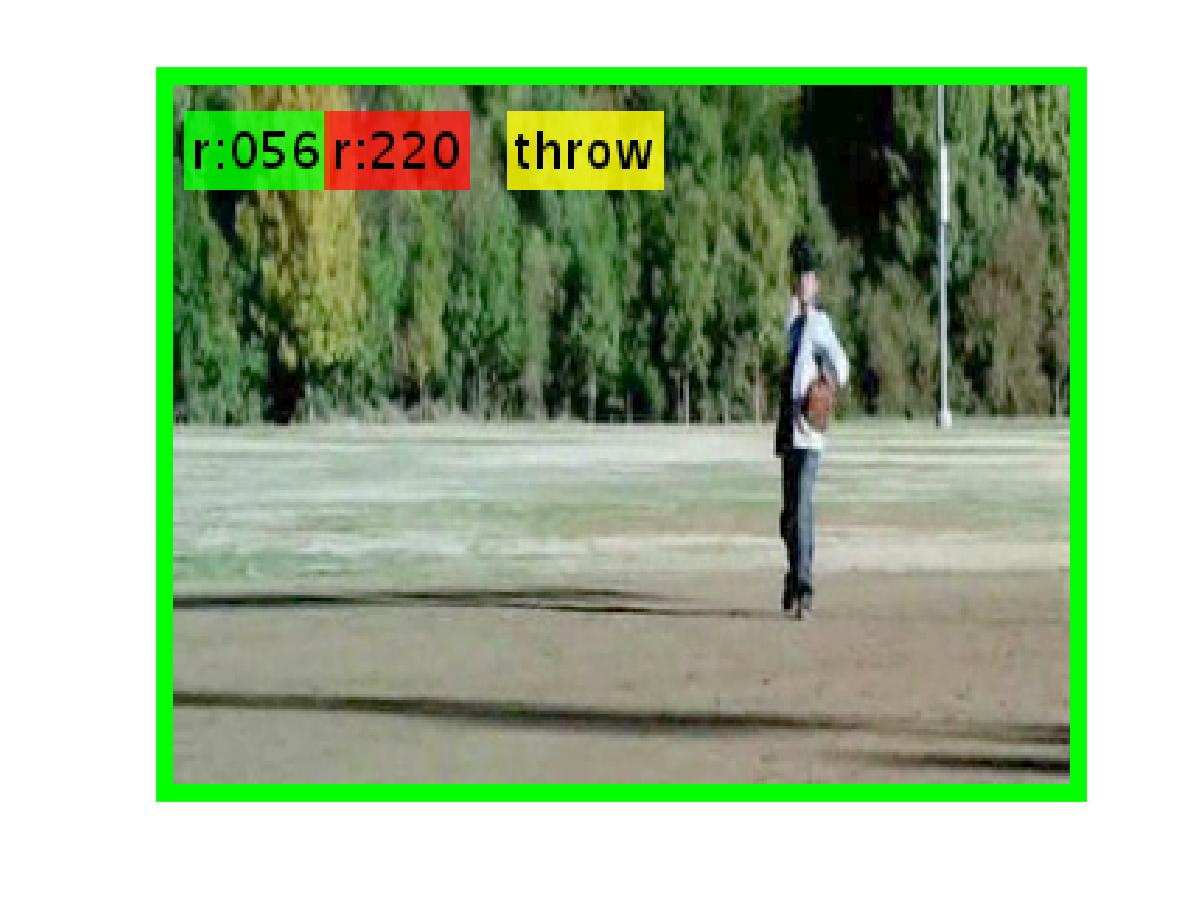}
\includegraphics[trim = 54mm 37mm 36mm 26mm, clip,scale=0.10]{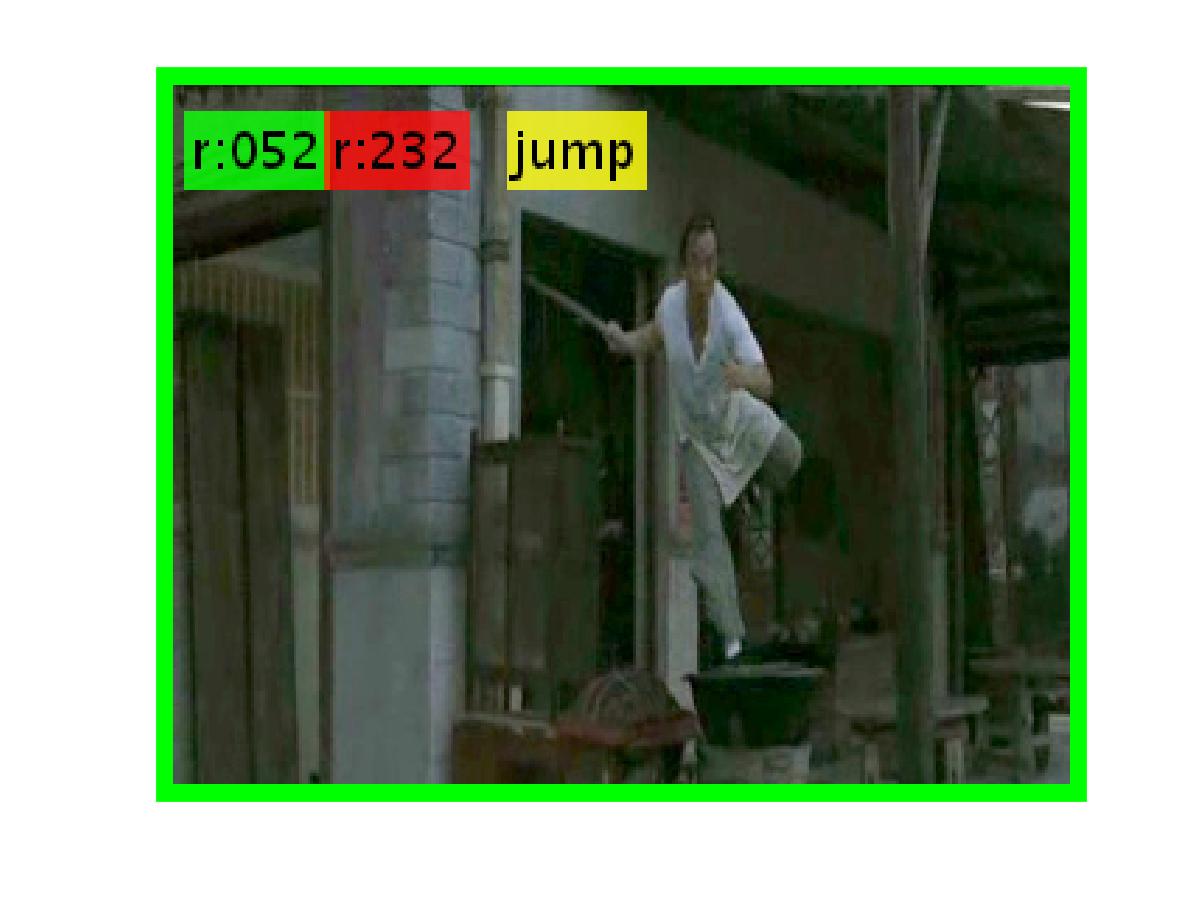}
\includegraphics[trim = 54mm 37mm 36mm 26mm, clip,scale=0.10]{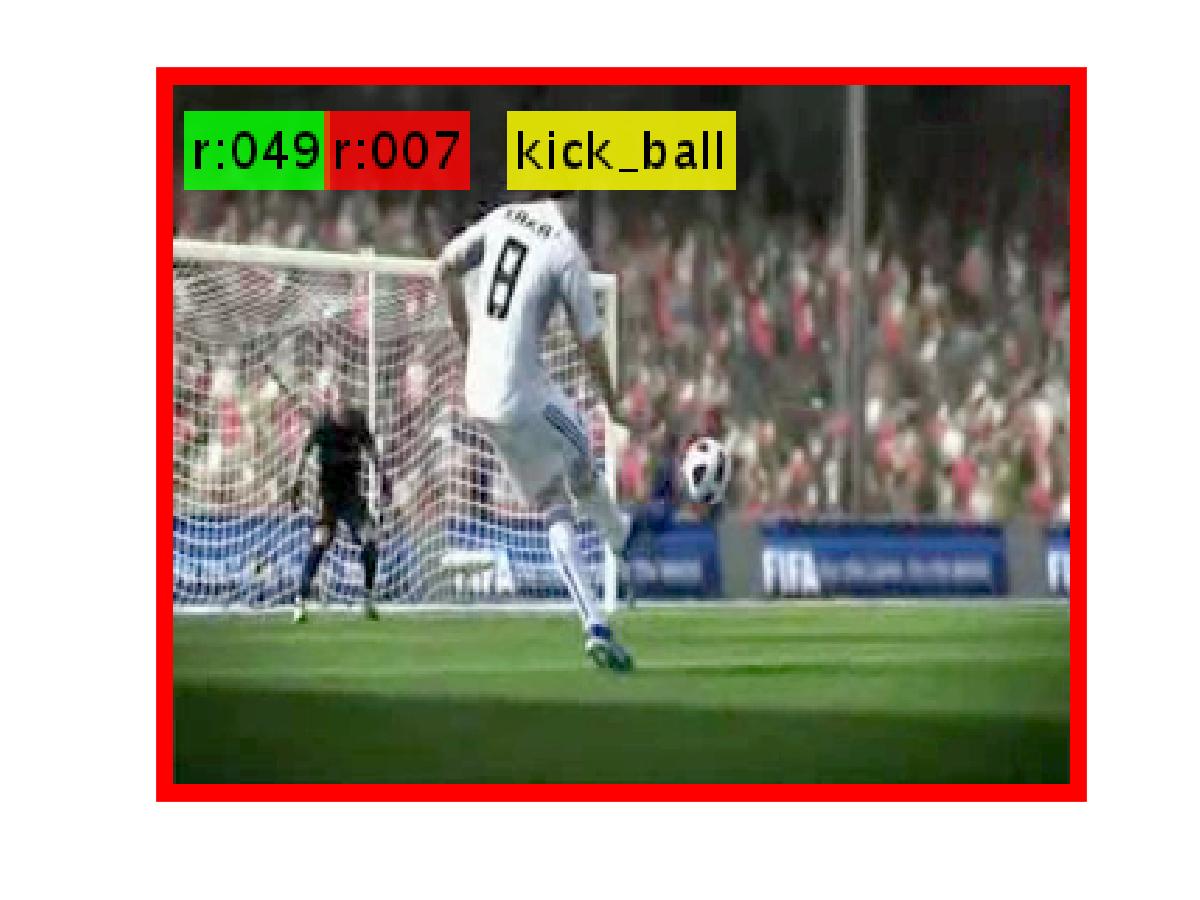}\\
\includegraphics[trim = 54mm 37mm 36mm 26mm, clip,scale=0.10]{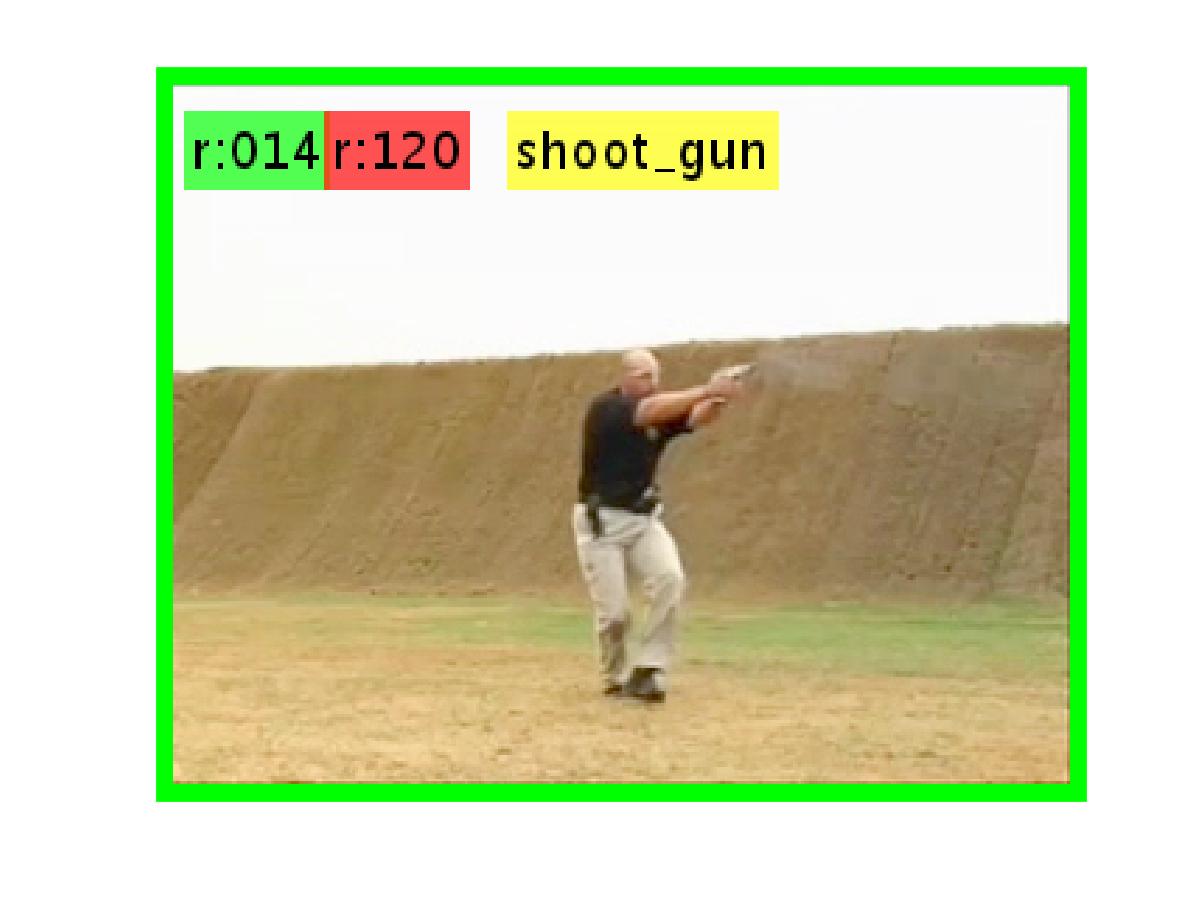}
\includegraphics[trim = 54mm 37mm 36mm 26mm, clip,scale=0.10]{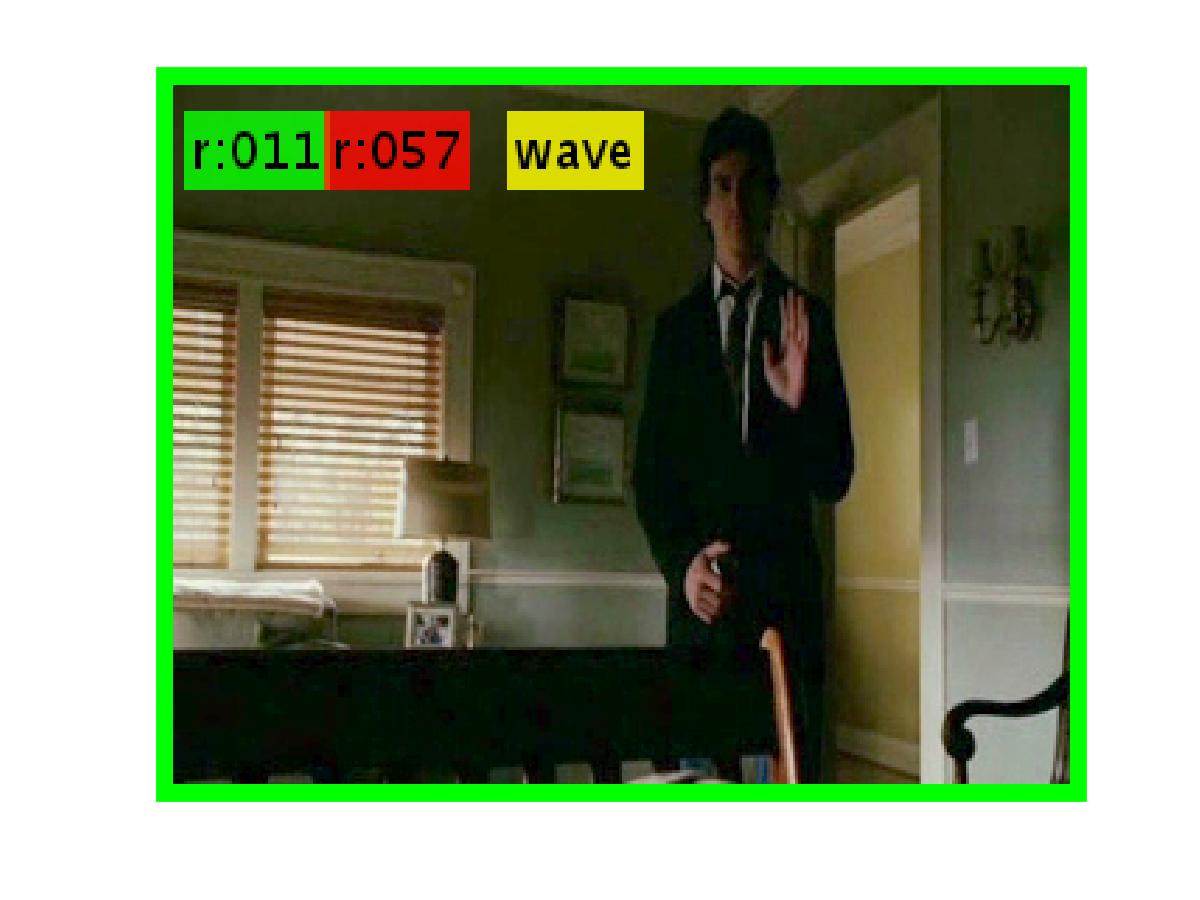}
\includegraphics[trim = 54mm 37mm 36mm 26mm, clip,scale=0.10]{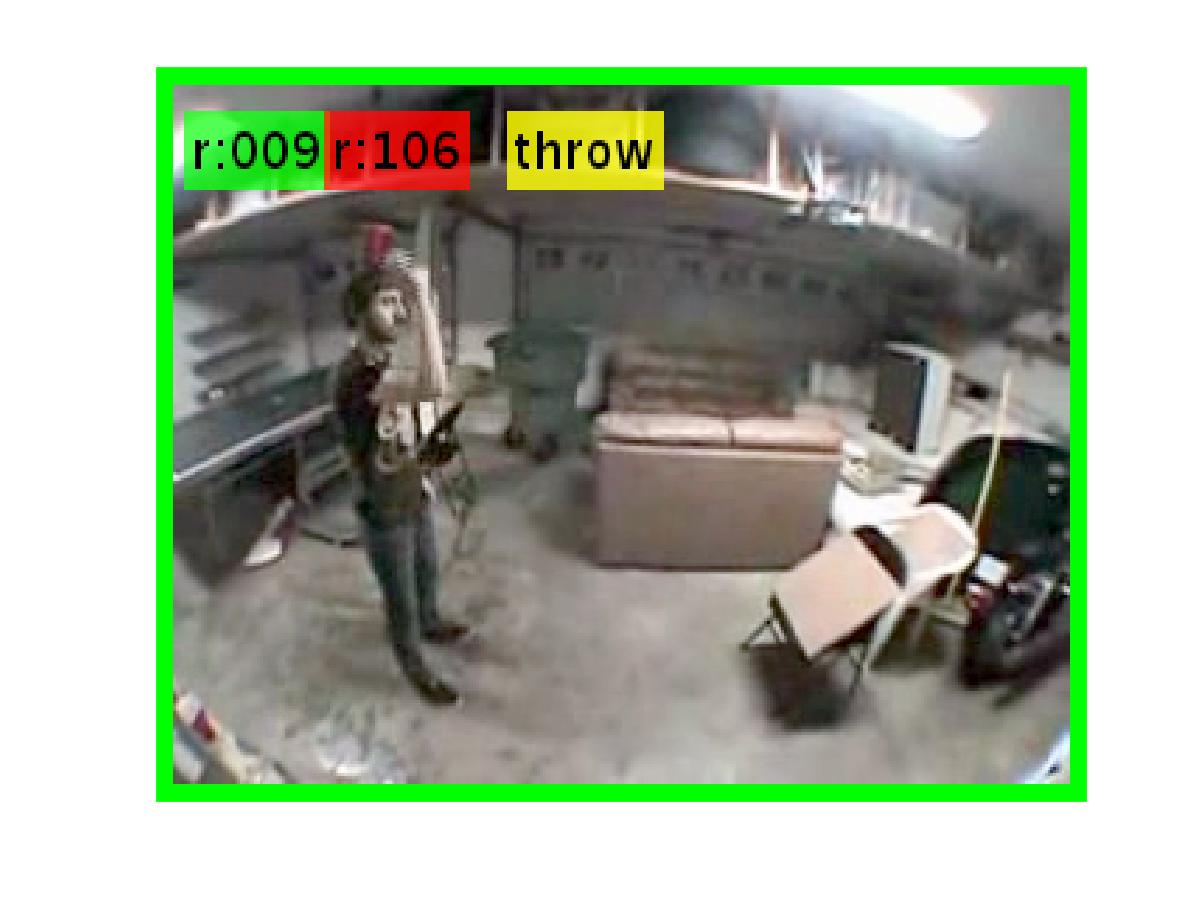}
\includegraphics[trim = 54mm 37mm 36mm 26mm, clip,scale=0.10]{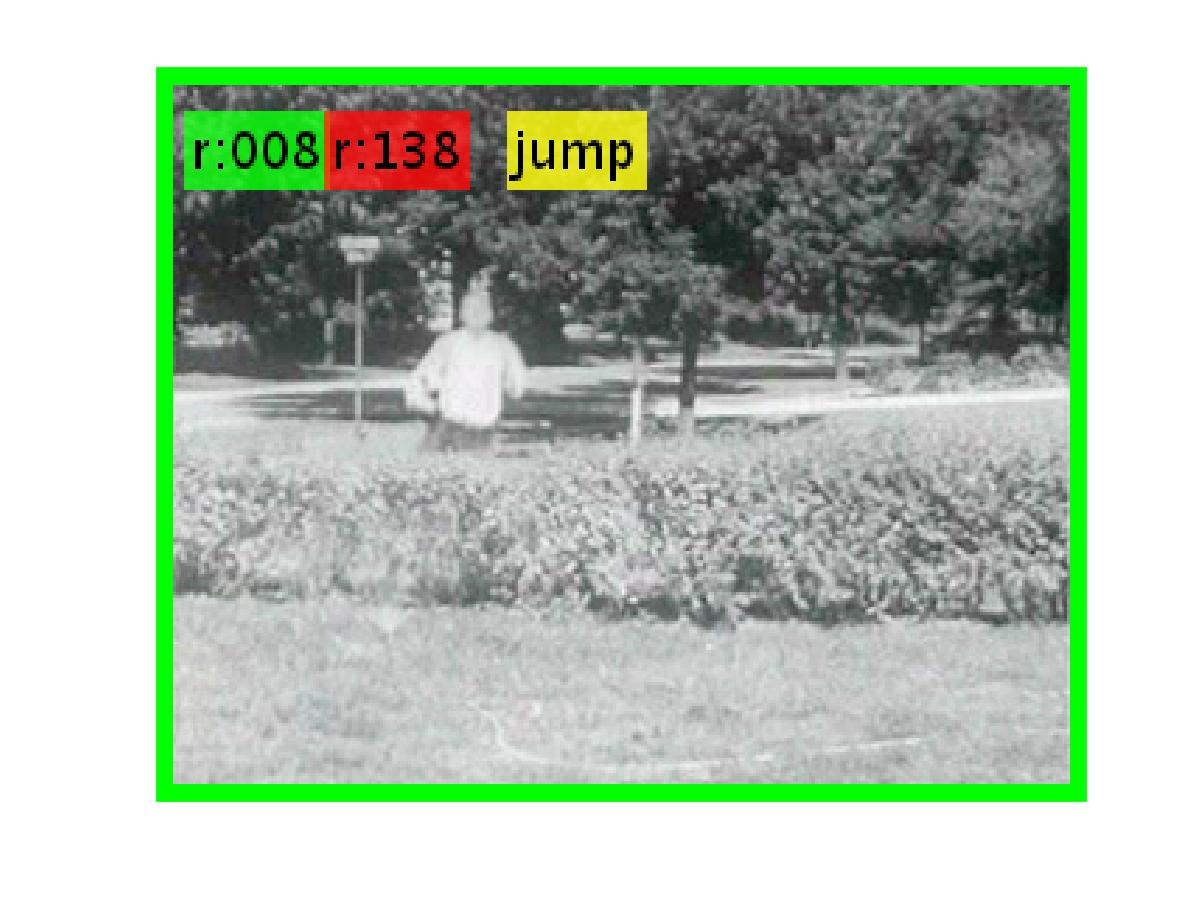}
\includegraphics[trim = 54mm 37mm 36mm 26mm, clip,scale=0.10]{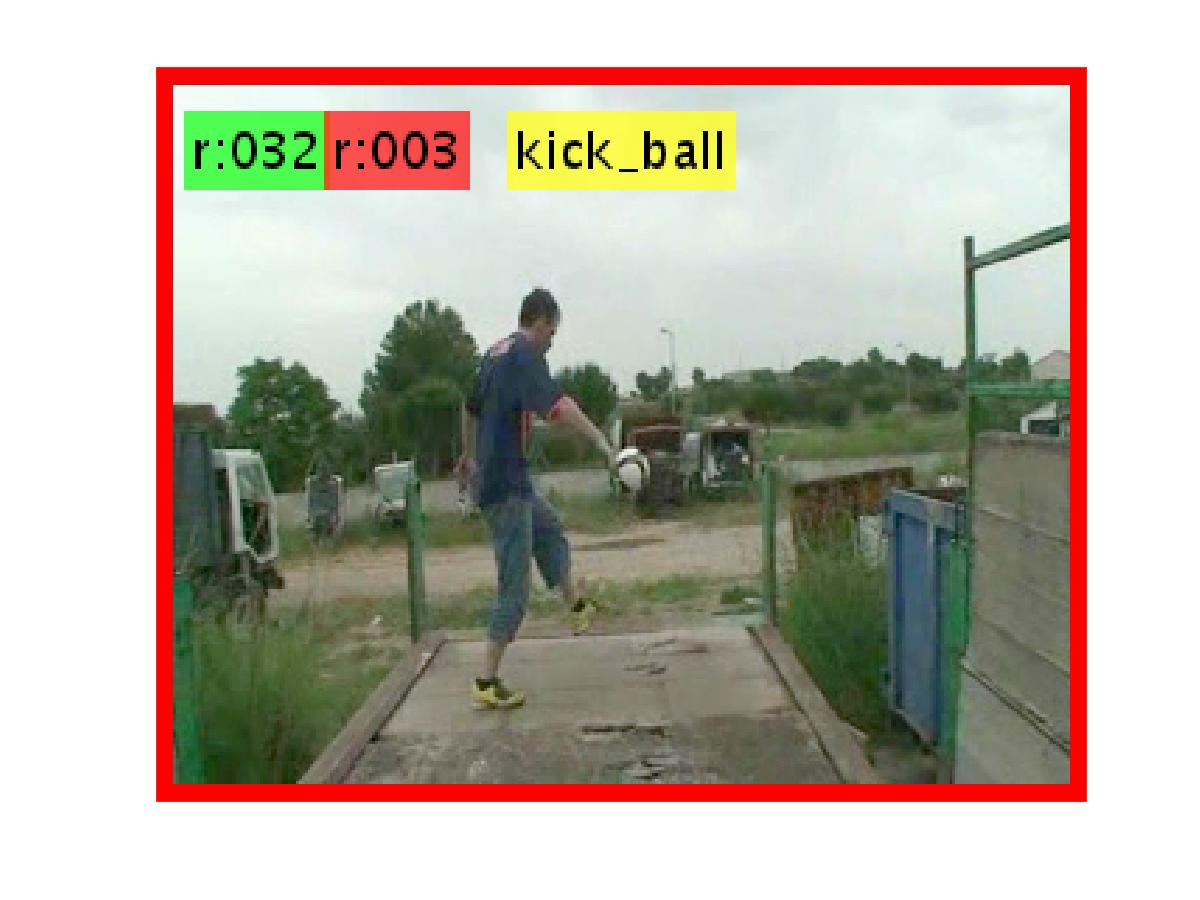}\vspace{-.3cm}\\
   \caption{Results on JHMDB-GT (split 1). Each column corresponds to an action class.
Video frames on the left (green) illustrate two test samples per action with the largest
improvement in ranking when using P-CNN (rank in green) and IDT-FV (rank in red).
Examples on the right (red) illustrate samples with the largest decreases in the ranking.
Actions with large differences in performance are selected according to Figure~\ref{fig:JHMDBbar}.
Each video sample is represented by its middle frame.
}
\label{fig:JHMDBqual}
\end{figure*}

Qualitative results comparing P-CNN and IDT-FV are presented in Figure~\ref{fig:JHMDBqual} for JHMDB-GT.
See Figure~\ref{fig:JHMDBbar} for the quantitative comparison.   
To highlight improvements achieved by the proposed
P-CNN descriptor, we show results for classes with a large
improvement of P-CNN over IDT-FV, such as \emph{shoot\_gun}, \emph{wave}, \emph{throw} and \emph{jump} as well
as for a class with a significant drop, namely \emph{kick\_ball}.
Figure~\ref{fig:JHMDBqual} shows two examples for each selected
action class with the maximum difference in ranks obtained by P-CNN
(green) and IDT-FV (red).
For example, the most significant improvement
(Figure~\ref{fig:JHMDBqual}, top left) increases the sample ranking 
from $211$ to $23$, when replacing IDT-FV by P-CNN. In particular, the \emph{shoot gun} 
and \emph{wave} classes involve small localized motion, making
classification difficult for IDT-FV while P-CNN benefits
from the local human body part information.
Similarly, the two samples from the action class \emph{throw}  also seem
to have restricted and localized motion while the action \emph{jump} is very
short in time. In the case of \emph{kick\_ball} the significant
decrease can be explained by the important dynamics of this action,
which is better captured by IDT-FV features. Note that P-CNN only
captures motion information between two consecutive frames. 

\begin{figure*}
\includegraphics[trim = 54mm 37mm 36mm 26mm, clip,scale=0.10]{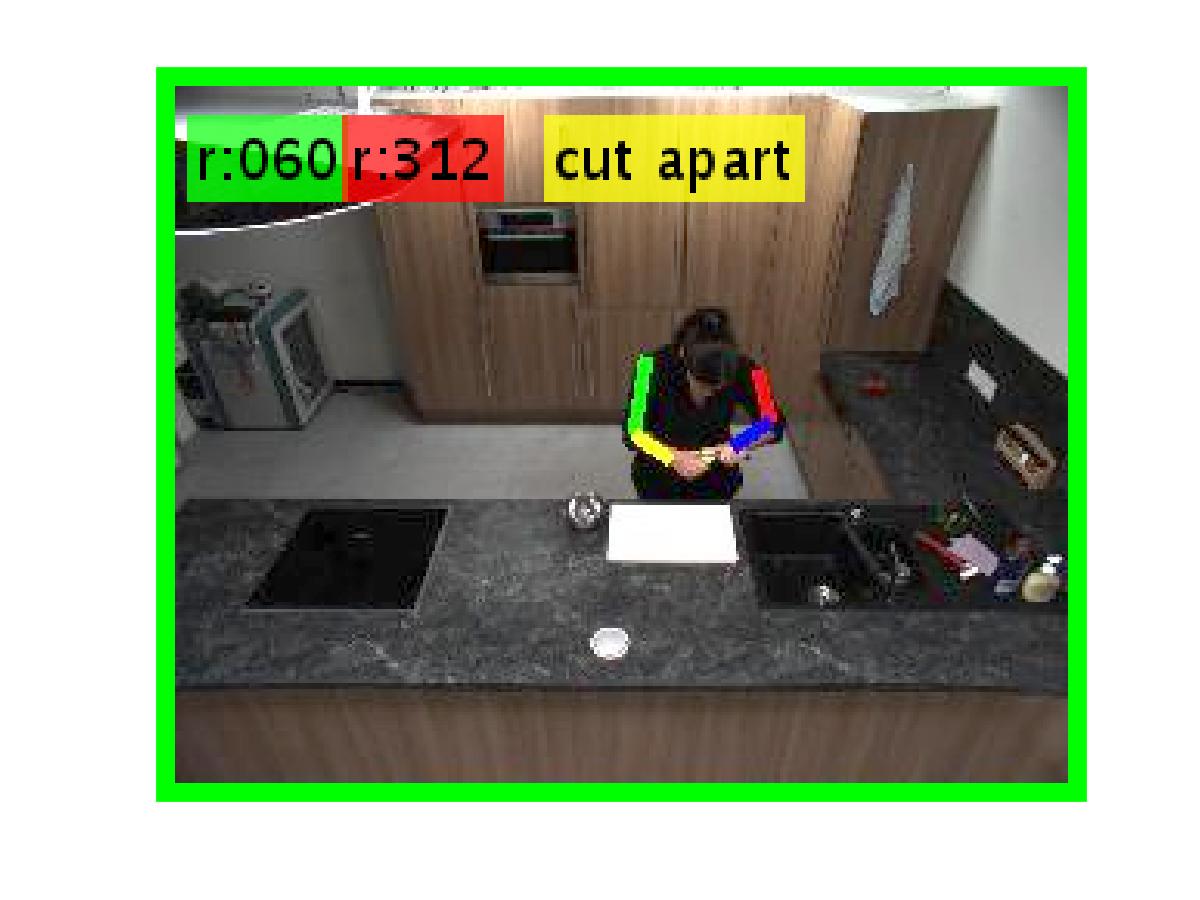}
\includegraphics[trim = 54mm 37mm 36mm 26mm, clip,scale=0.10]{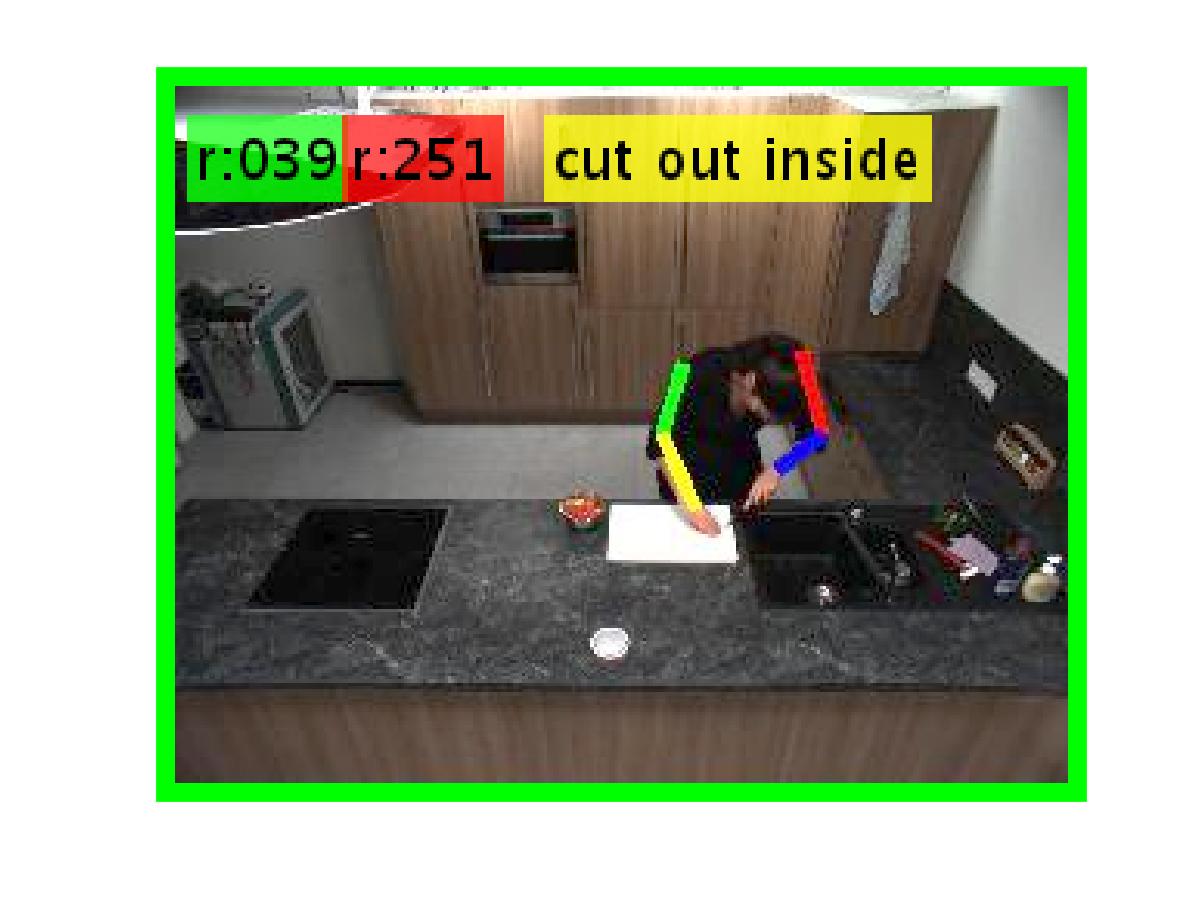}
\includegraphics[trim = 54mm 37mm 36mm 26mm, clip,scale=0.10]{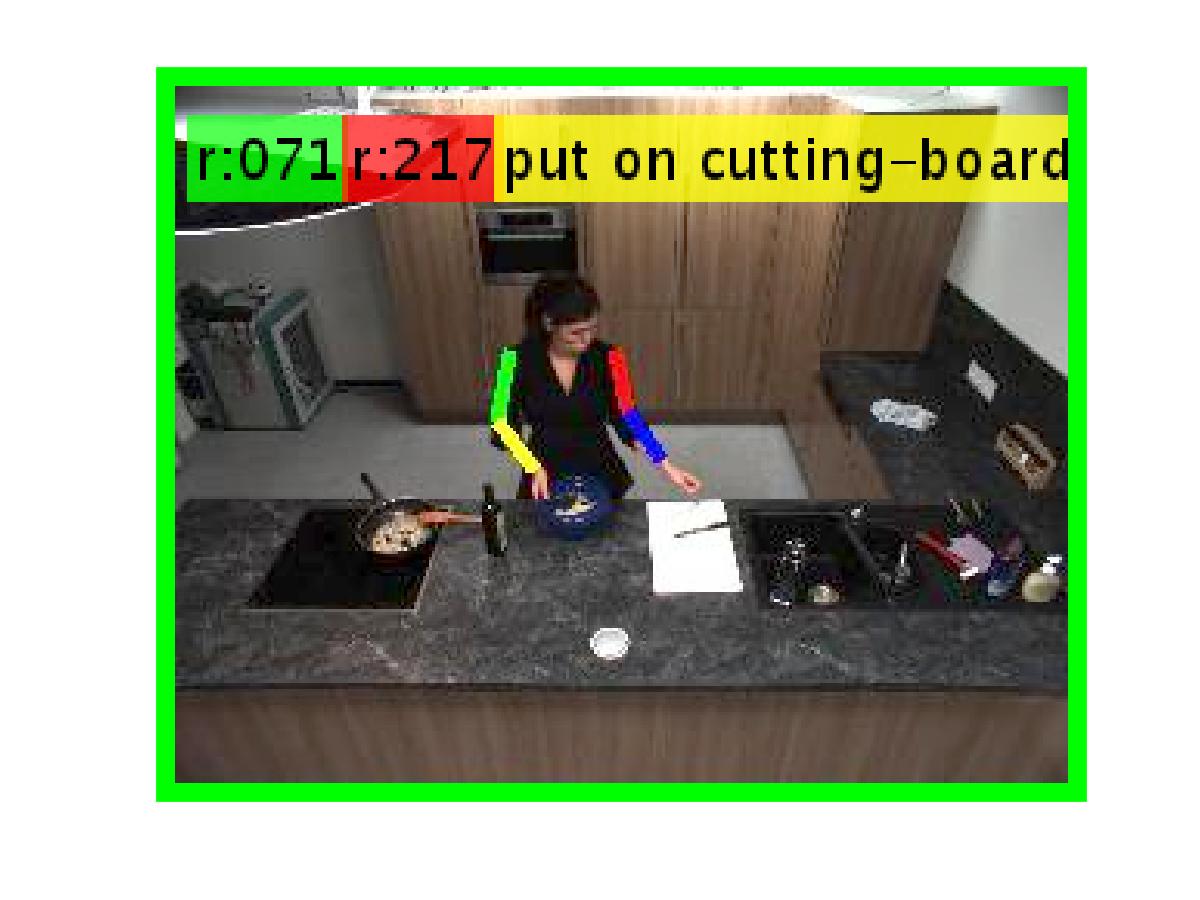}
\includegraphics[trim = 54mm 37mm 36mm 26mm, clip,scale=0.10]{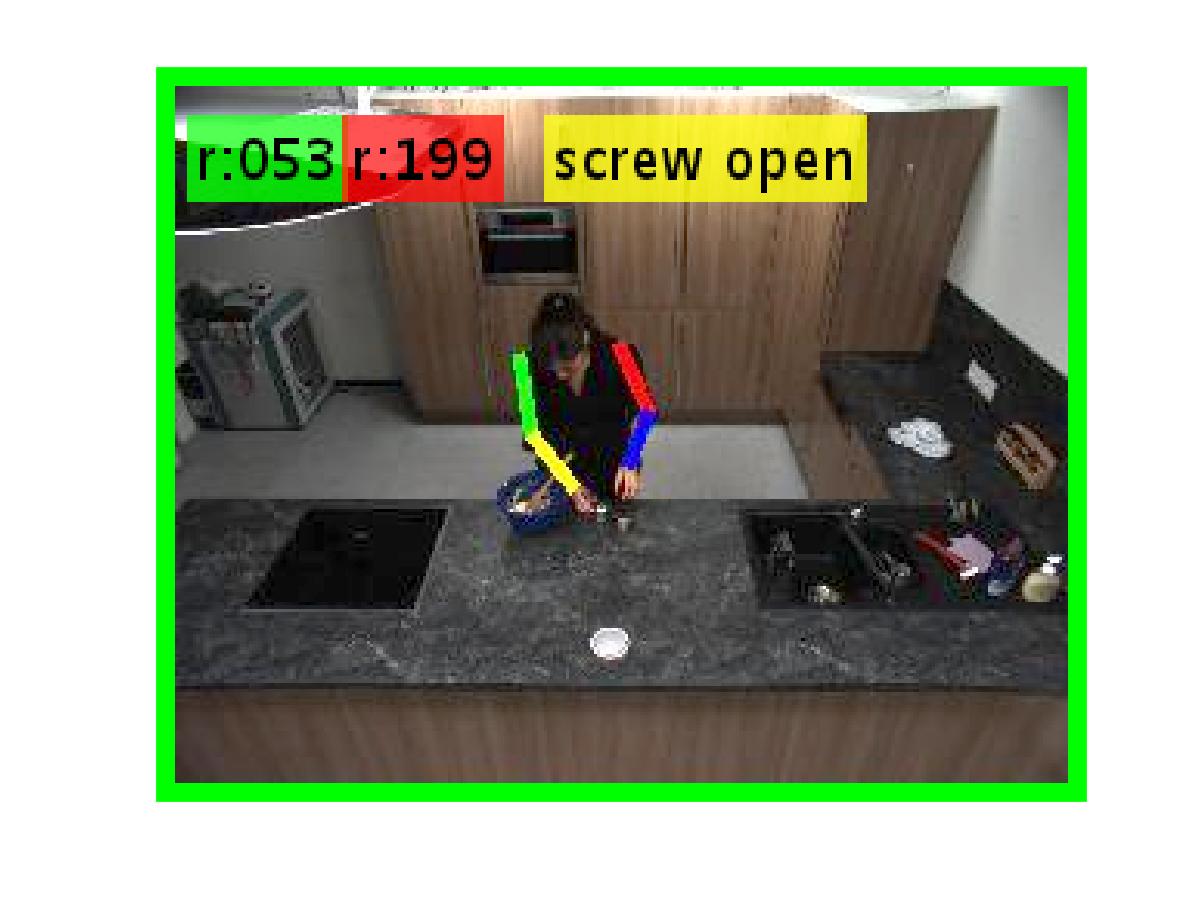}
\includegraphics[trim = 54mm 37mm 36mm 26mm, clip,scale=0.10]{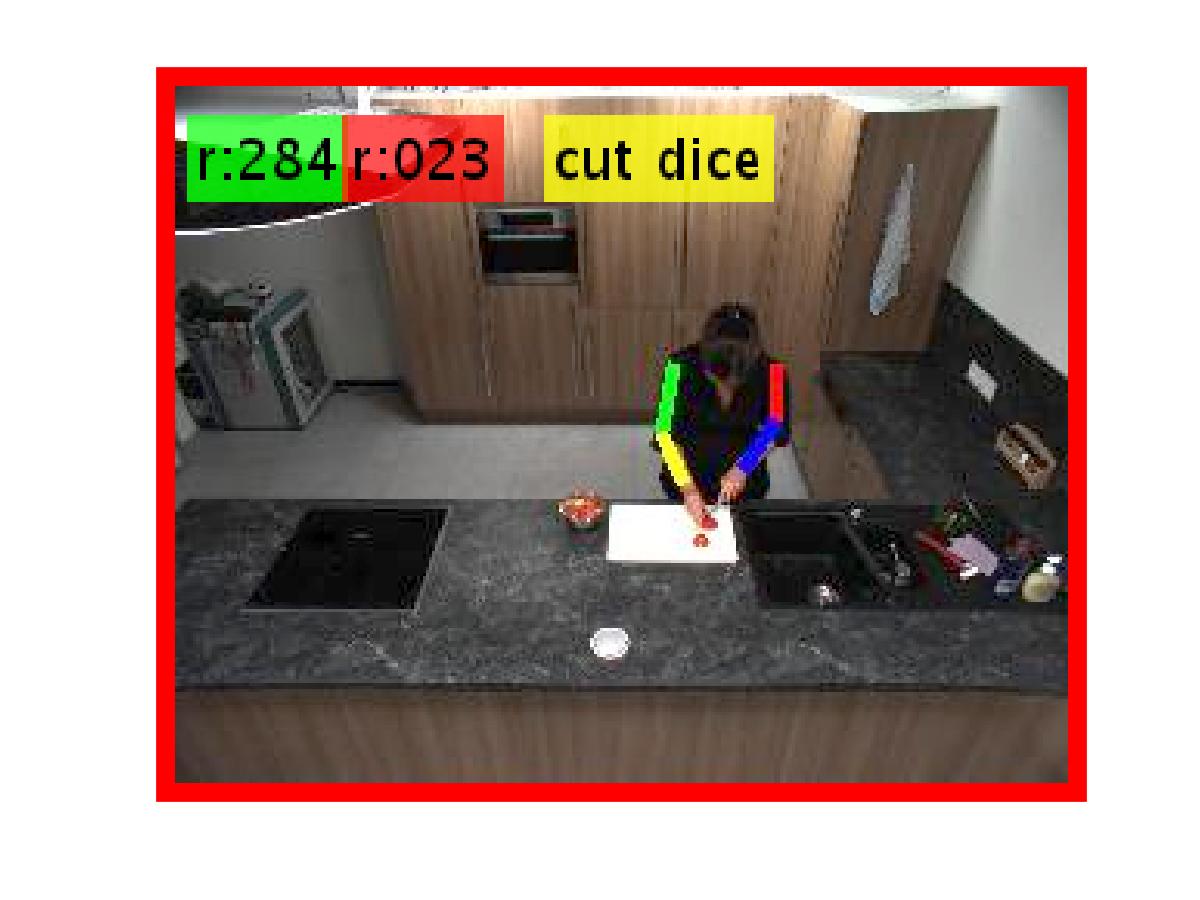}\\
\includegraphics[trim = 54mm 37mm 36mm 26mm, clip,scale=0.10]{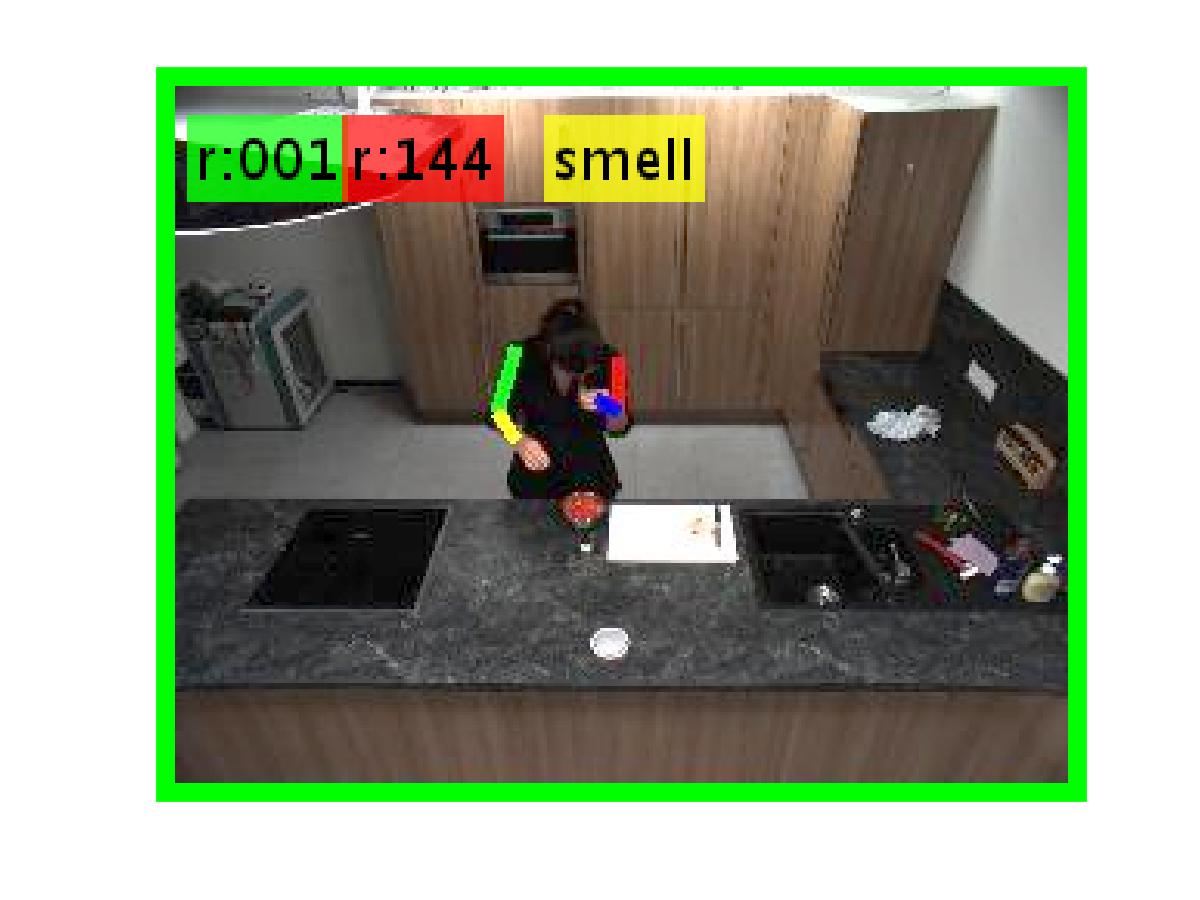}
\includegraphics[trim = 54mm 37mm 36mm 26mm, clip,scale=0.10]{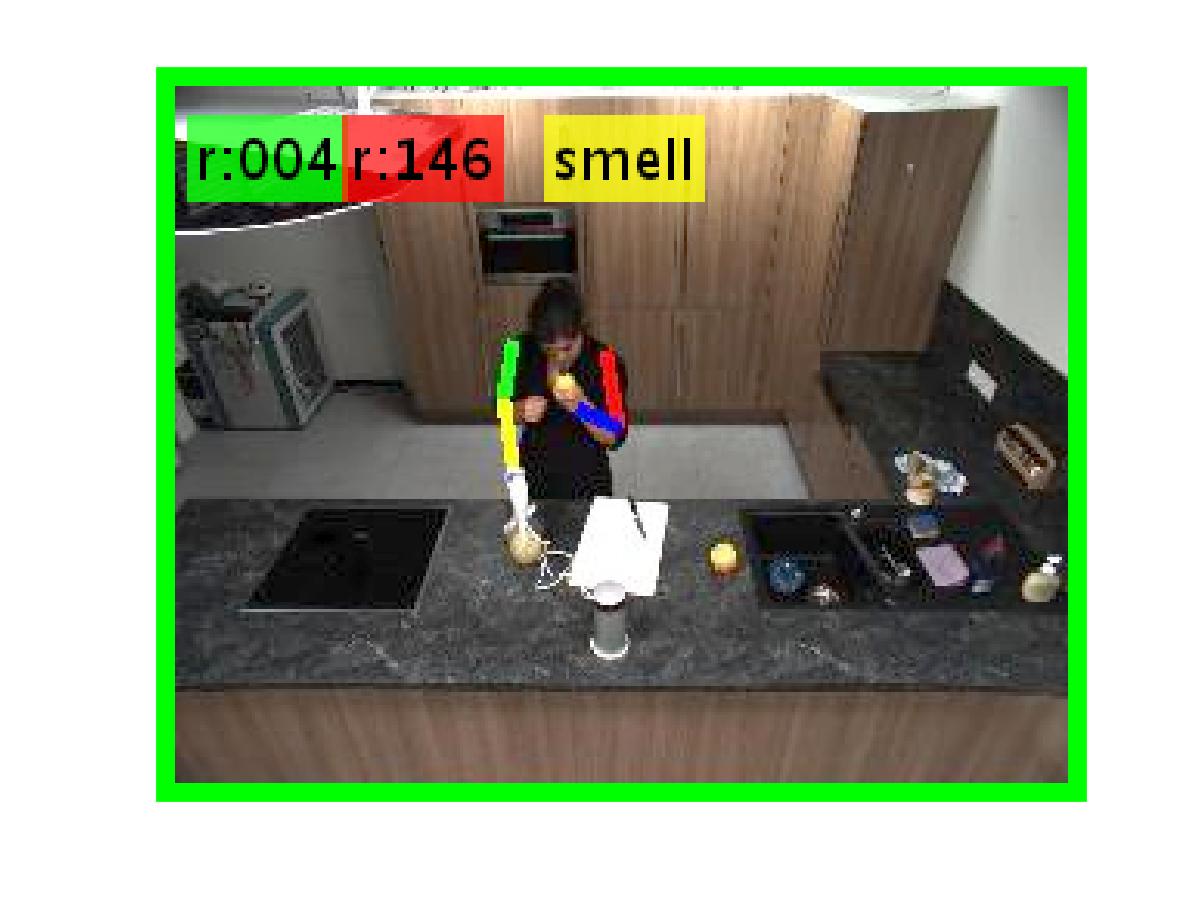}
\includegraphics[trim = 54mm 37mm 36mm 26mm, clip,scale=0.10]{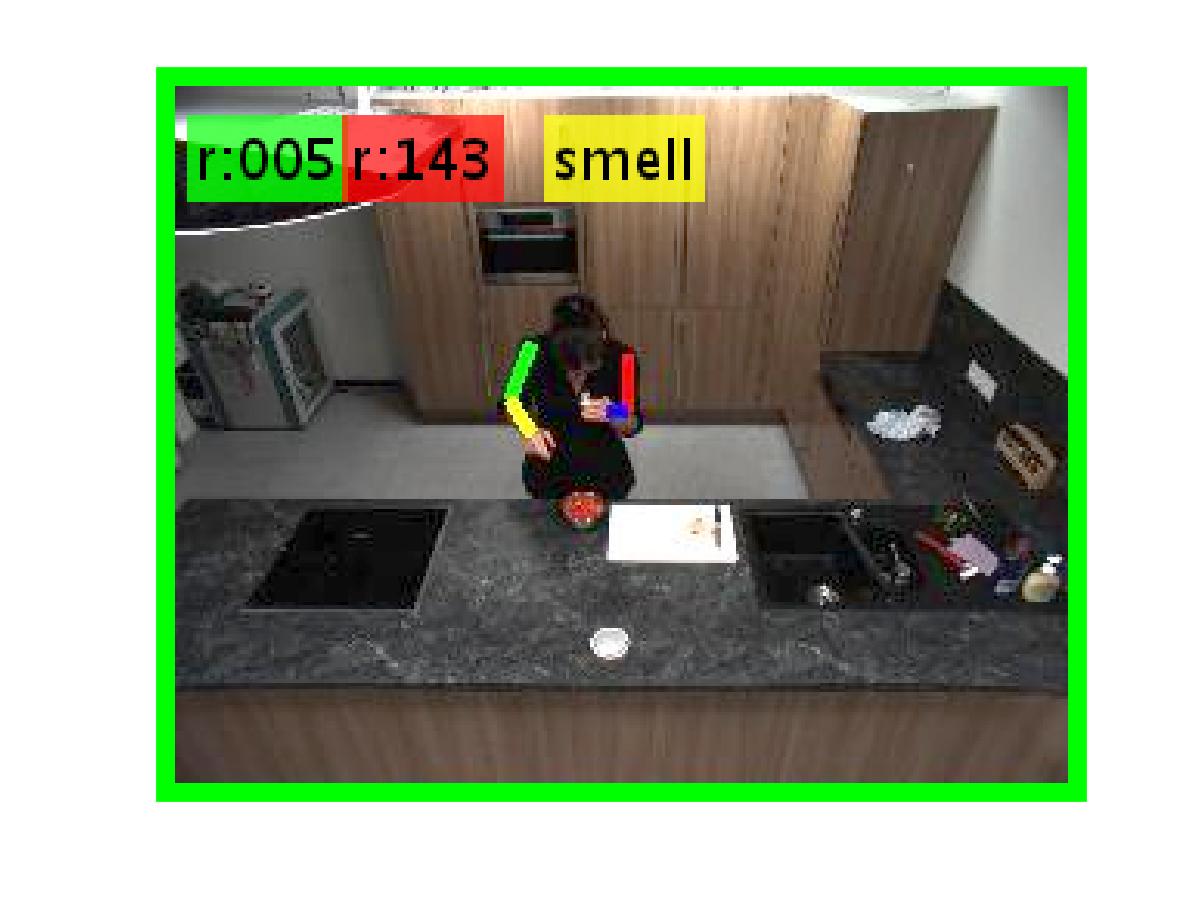}
\includegraphics[trim = 54mm 37mm 36mm 26mm, clip,scale=0.10]{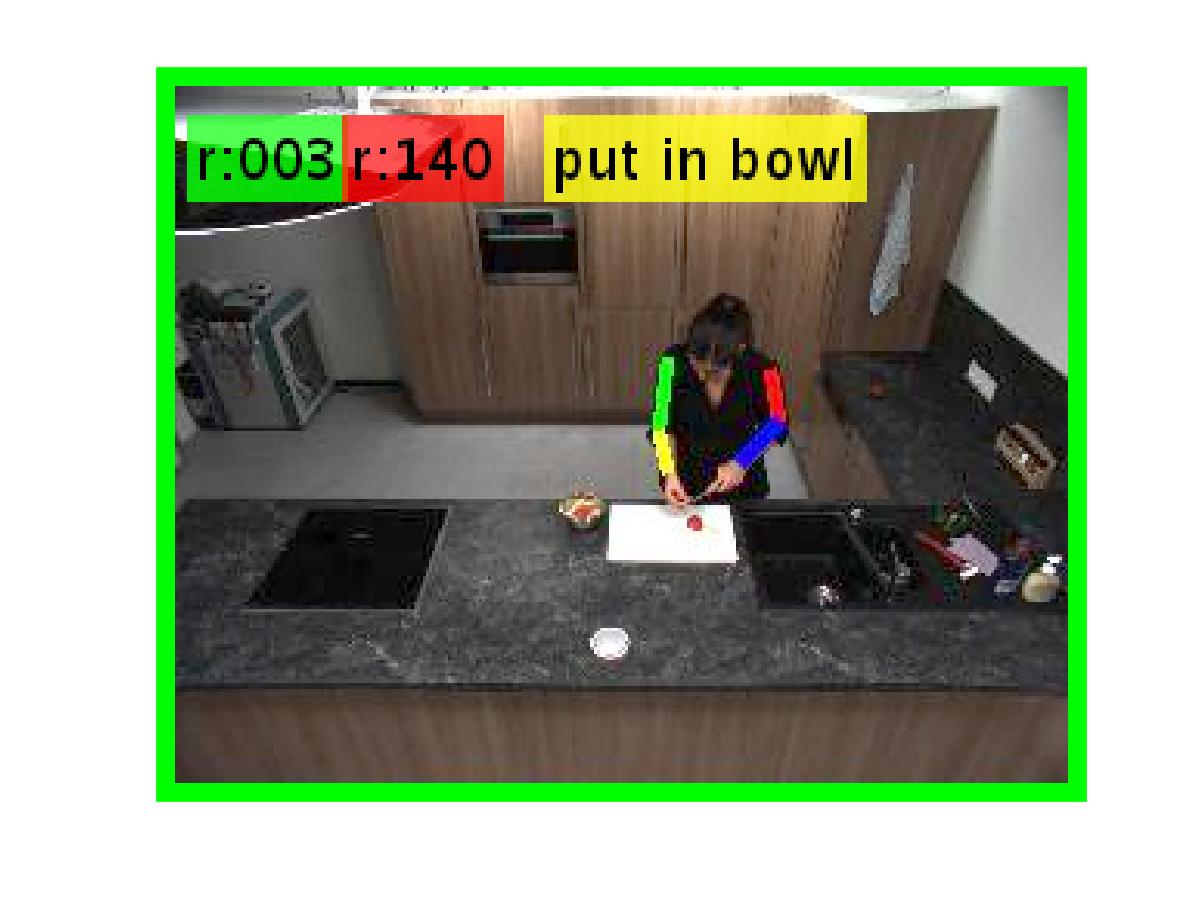}
\includegraphics[trim = 54mm 37mm 36mm 26mm, clip,scale=0.10]{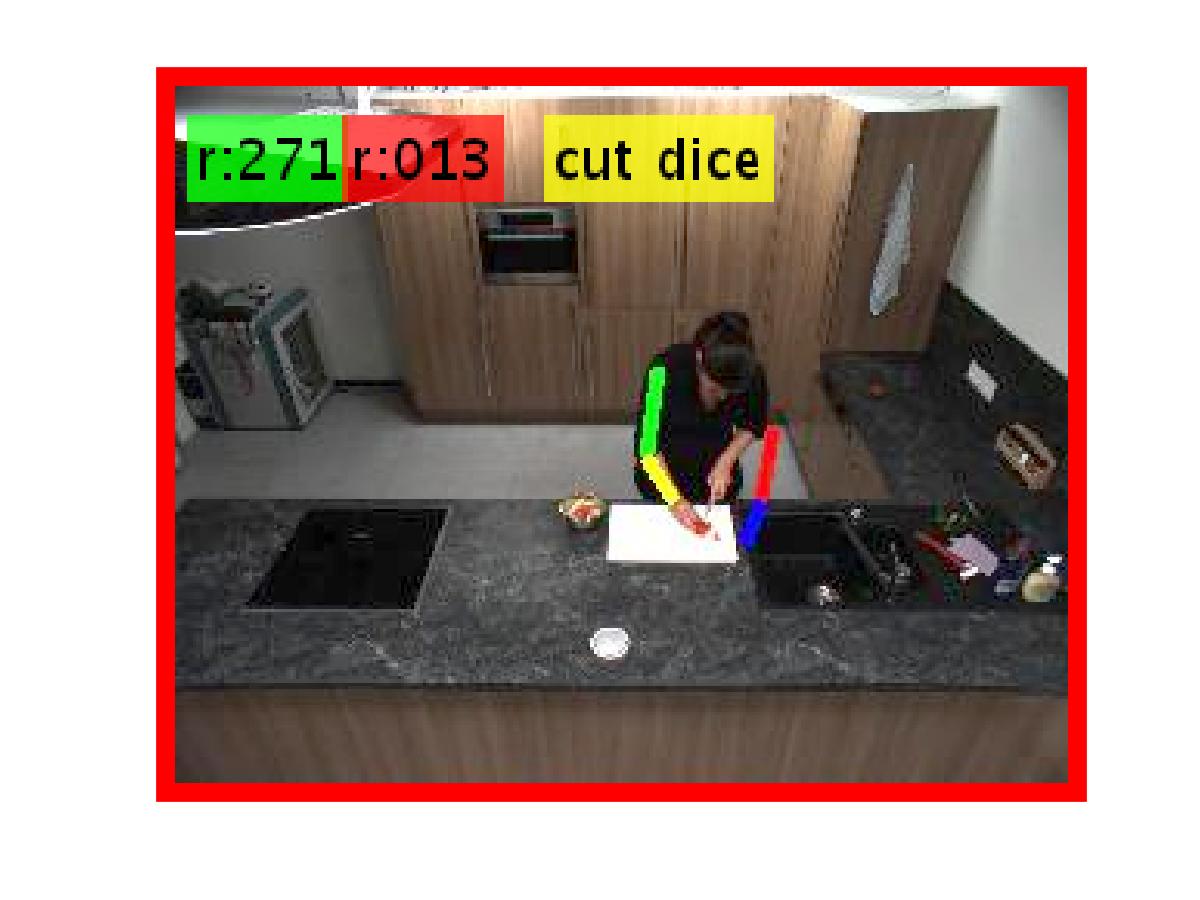}\vspace{-.3cm}\\
\caption{Results on MPII Cooking-Pose~\cite{Cherian14} (split 1).
Examples on the left (green) show the $8$ best ranking improvements (over all classes) obtained by using P-CNN (rank in green) instead of IDT-FV (rank in red).
Examples on the right (red) illustrate video samples with the largest decrease in the ranking. Each video sample is represented by its middle frame.
}
\label{fig:MPIIqual}
\end{figure*}

Figure~\ref{fig:MPIIqual} presents qualitative results for MPII
Cooking-Pose~\cite{Cherian14} showing samples with the maximum
difference in ranks over all classes. 

\begin{table}\centering
\renewcommand{\arraystretch}{0.9}
 \begin{tabular}{@{}lc@{}}\toprule
    \textbf{Method}                                                            & \textbf{MPII Cook.}  \\ \midrule
	Holistic + Pose~\cite{rohrbach12cvpr} 		                                   & $57.9$  \\
	Semantic Features~\cite{zhou14eccv}                                               & $70.5$  \\
        Interaction Part Mining~\cite{zhou2015interaction}       & $72.4$ \\
	P-CNN + IDT-FV (our)     	                                         & $71.4$  \\
	\bottomrule 
 \end{tabular}
  \caption{State of the art on the MPII Cooking (\% mAP).}
 \label{globresults}
\end{table}

\section{Conclusion}
\label{conclusion}
This paper introduces  pose-based convolutional neural network features
(P-CNN). Appearance and flow information is extracted at characteristic  
positions obtained from human pose and aggregated over frames of a
video. Our P-CNN description is shown to be significantly  more robust to
errors in human pose estimation compared to existing pose-based
features such as HLPF~\cite{jhuang:hal-00906902}.
In particular, P-CNN significantly outperforms HLPF on the task
of fine-grained action recognition in the MPII Cooking Activities dataset.
Furthermore, P-CNN features are complementary to 
the dense trajectory features and significantly improve the current
state of the art for action recognition when combined with IDT-FV.  

Our study confirms conclusions in~\cite{jhuang:hal-00906902}, namely, that
correct estimation of human poses leads to significant improvements in action recognition.
This implies that pose is crucial to capture discriminative information of human actions.
Pose-based action recognition methods have a promising future due to
the recent progress in pose estimation, notably using CNNs~\cite{Chen_NIPS14}.
An interesting direction for future work is to adapt 
CNNs for each P-CNN part (hands, upper body, etc.) by
fine-tuning networks for corresponding image areas.
Another promising direction is to model temporal evolution of frames
using RNNs~\cite{donahue2015,rumelhart1986learning}.

\section*{Acknowledgements}
This work was supported by the MSR-Inria joint
lab, a Google research award, the ERC starting grant ACTIVIA and the ERC advanced grant ALLEGRO.

{\small
\bibliographystyle{ieee}
\bibliography{shortstrings,egbib}
}

\end{document}